\documentclass[10 pt,journal]{IEEEtran}  	% USLetter
\IEEEoverridecommandlockouts	% to be used with the \thanks command
\usepackage{graphics}		% for pdf, bitmapped graphics files
\usepackage{epsfig}			% for postscript graphics files
\usepackage{amsmath}		% assumes amsmath package installed
\usepackage{amssymb}		% assumes amsmath package installed

\usepackage{hyperref}
\usepackage{algorithm}		% for IEEE-formatted algorithms description
\usepackage{algpseudocode}	% for IEEE-formatted pseudo-code usage in algorithms
\usepackage[table]{xcolor}
\usepackage{lipsum}
\usepackage{array}
\usepackage{float}
\usepackage{color}
\usepackage{booktabs}
\usepackage{bm}
\usepackage{subcaption}
\graphicspath{ {fig/} }
\usepackage{xr}
\usepackage[normalem]{ulem}

\usepackage{epstopdf}
\usepackage{color}

\newcolumntype{C}[1]{>{\centering\let\newline\\\arraybackslash\hspace{0pt}}m{#1}}

\usepackage{soul}

\title{\LARGE \bf A Hierarchical Architecture for Human-Robot Cooperation Processes}
\author{
Kourosh Darvish$^{1}$, Enrico Simetti, Fulvio Mastrogiovanni, Giuseppe Casalino
\thanks{
All the authors are with the Department of Informatics, Bioengineering, Robotics, and Systems Engineering, University of Genoa, Via Opera Pia 13, 16145, Genoa, Italy.}
\thanks{This paper has been accepted for publication in the IEEE Transactions on Robotics (T-RO).}
\thanks{$^{1}$Corresponding author's email: kourosh.darvish@gmail.com}}%
\usepackage{tikz}
\usepackage{textcomp}
\usepackage{hyperref}
\usepackage{lipsum}

\newcommand\copyrighttext{%
  \footnotesize \textcopyright 2020 IEEE. Personal use of this material is permitted.
  Permission from IEEE must be obtained for all other uses, in any current or future
  media, including reprinting/republishing this material for advertising or promotional
  purposes, creating new collective works, for resale or redistribution to servers or
  lists, or reuse of any copyrighted component of this work in other works.
  }
\newcommand\copyrightnotice{%
\begin{tikzpicture}[remember picture,overlay]
\node[anchor=south,yshift=5pt] at (current page.south) {\fbox{\parbox{\dimexpr\textwidth-\fboxsep-\fboxrule\relax}{\copyrighttext}}};
\end{tikzpicture}%
}

\begin{document}

\maketitle

\copyrightnotice

\thispagestyle{empty}
\pagestyle{empty}

\begin{abstract}
%A flexible assembly of semi-finished products is one of the expected features of next-generation shop-floor lines. Currently, such flexibility is placed on the shoulders of human operators.

%At the same time, operations in the shop-floor are still structured and well-defined. Collaborative robots have been designed to allow for a transition of such a burden from human operators to robots.

In this paper we propose \textsc{FlexHRC+}, a hierarchical human-robot cooperation architecture designed to provide collaborative robots with an extended degree of autonomy when supporting human operators in high-variability shop-floor tasks.
%, such as the ones expected to occur in next-generation shop-floor lines.
The architecture encompasses three levels, namely for perception, representation, and action.
% each one aimed at addressing specific traits of human-robot cooperation processes.
% \textcolor{blue}{\sout{Starting from the architecture introduced in \cite{Darvish2018Mechatronics},}} h
Building up on previous work, here we focus on
(i) an in-the-loop decision making process for the operations of collaborative robots coping with the variability of actions carried out by human operators, and 
(ii) the representation level, integrating a hierarchical AND/OR graph whose online behaviour is formally specified using First Order Logic.
The architecture is accompanied by experiments including collaborative furniture assembly and object positioning tasks.
% The behaviour of \textsc{FlexHRC+} is discussed using three experiments related to collaborative table assembly and object positioning tasks involving intra-task and inter-tasks variability.
\end{abstract}


%%%%%%%%%%%%%%%%%%%%%%%%%%%%%%%%%%%%%%%%%%%%%%%%%%%%%%%%%%%%%%%%%%%%%%%%%%%%%%%

\section{Introduction}
\label{sec:Introduction}
% \textbf{[1 page]}

% \textcolor{blue}{ -- An introduction to HRC, what are the needs in an unstructured environment and dynamic environment : }

The paradigm of consumer- and demand-driven manufacturing introduces the need for small-scale, customised, high quality production at lower prices, and with faster delivery times \cite{SRA2014, Esmaeilian2016}.
%The paradigm targets small-scale production, which is typical of medium-size manufacturing enterprises (SMEs).
However, small-scale production does not fully exploit the benefit of robot-based manufacturing yet \cite{Kock2011, MAR2017}.
%High wage costs cause small-scale manufacturing to be delocalised to low wage countries \cite{Kock2011, SRA2014}.  
%According to the International Federation of Robotics (IFR), only 10\% of manufacturing jobs currently achieve full automation \cite{IFR2018}.
Consumer- and demand-driven manufacturing requires robots characterised by high flexibility, fast reconfiguration and installation, as well as low maintenance costs.
Collaborative robots are expected to meet such demands, decrease costs, and therefore increase products variability and customisation \cite{SRA2014, Esmaeilian2016, Kock2011}.
In fact, they are considered key enabling factors to automate small-scale production when operations to be carried out are highly dynamic and partially unstructured \cite{IFR2018}.

% Combining the humans and safe robots individual capabilities together to cooperate in a dynamic and partially-structured environments is an approach to reach the features of the new manufacturing paradigms. 

% For an effective and efficient cooperation the robotic system should enable the three main characteristics of the manufacturing paradigms.

% should adapt to the new manufacturing paradigms where flexible and agile production and frequent reconfiguration is necessary, such that the \textit{return-on-investment} increases. 

% In the new production scenarios, the assembly process should not dedicate to a particular product but instead should be reconfigurable quickly for large product variety. Till now mainly high volume productions benefits from fully automation line with quick and accurate and costly robots which needs high initial investment, whereas the small or medium size productions are done manually \cite{Esmaeilian2016, Kock2011, Lasi2014}.

% We believe that to respond the smart factories requirements, the robotic system should be flexible, scalable, and safe as well as the efficient and effective in the smart factories. With these features, they reduce the costs and enhance the delivery time, quality, and customization. 

% \textbf{Safety and human factors}

Recently, many authors argued that consumer- and demand-driven manufacturing can benefit from the introduction of human-robot cooperation (HRC) processes \cite{Cherubinietal2016}. 
HRC assumes that human operators and robots purposely interact in a shared workspace to achieve a common objective.
The design of collaborative robots should adhere to a number of \textit{human-centric} principles.
%so that the cooperation can be effective, efficient, and natural.
Human-centric design enforces such factors as the \textit{explainability} of robot decisions, the \textit{usability} of robot interfaces, the \textit{awareness} of the cooperation process, and a fair human-robot \textit{workload}, as well as \textit{safety} requirements for human operators \cite{Adams2005, Steinfeld2006}. 
On the one hand, since it has been shown that the \textit{effectiveness} and the overall performance of human operators are positively correlated with robot motion predictability \cite{Bortot2013}, collaborative robots should 
%be characterised by low payload and low inertia, human-friendly mechanical design, power and speed limitations, and safe-by-design collision detection.
%Furthermore, collaborative robots should 
(requirement $R_1$) prevent psychological discomfort, stress, and a high induced cognitive load on human operators \cite{DeSantis2008, Kock2011, Lasota2017}. 
On the other hand, it has been demonstrated that a natural and efficient cooperation is possible only by a reasoned trade-off between the cooperation objective (e.g., the assemblage of a semi-finished product), and the human or robot degrees of autonomy.
This is specially true when the task is only partially well-defined (e.g., such assemblage can be done using different action sequences), which can be somewhat enforced or relaxed on a context-dependent basis \cite{Goodrich2007, Ferland2013}.
Collaborative robots should ($R_2$) be able to react to human operator actions while retaining the capability of planning goal-oriented action sequences \cite{Valli2008, Argall2009, Darvish2018Mechatronics}, and ($R_3$) doing so by abstracting the \textit{structure} of tasks from perceptual variability and uncertainties \cite{DeSantis2008, Kock2011, Darvish2018Roman}. 
The customised production advocated by consumer- and demand-driven manufacturing still relies on the cognitive capabilities of human operators since collaborative robots are largely unable to \textit{efficiently} manage inter-tasks or intra-task variations \cite{IFR2018}.
The ability of human operators to decompose complex tasks into simpler operations (e.g., assembling furniture parts to obtain other semi-finished parts to be used later), or to \textit{naturally} manage small variations (e.g., assembling furniture with parts of different size, like tables with differing flat top size or leg length), still poses a significant challenge for collaborative robots \cite{Garcia2013}.
As a consequence, collaborative robots ($R_4$) should exhibit decision making capabilities grounded on flexible task representations, and ($R_5$) should enforce a definition of hierarchical action sequences able to map high-level complex tasks to low-level, simple robot operations.

In this paper, we present an integrated architecture for HRC processes, which we refer to as \textsc{FlexHRC+}, aimed at addressing requirements $R_2$, $R_4$ and $R_5$ outlined above. 
\textsc{FlexHRC+} can adapt the behaviour of collaborative robots to human operator actions, while proactively taking decisions aimed at meeting the cooperation goals (addressing $R_2$).
\textsc{FlexHRC+} enables online human-robot decision making, a flexible execution of HRC tasks (addressing $R_4$), and a hierarchical representation of such tasks enforcing modularity and reuse (addressing $R_5$).
% \textcolor{blue}{-- Novelties of the paper:}
Our contribution is at two levels.
\begin{itemize}
\item
\textit{Human-robot cooperation level}. 
The first contribution is an in-the-loop, hybrid reactive-deliberative architecture for online, flexible and scalable HRC processes. 
The architecture is characterised by proactive decision making and reactive adaptation to a perceived sequence of human operator actions or unsuccessful robot operations. 
%\textsc{FlexHRC+} enables flexibility at the task and human-robot team levels: while the former enables human-robot cooperation process with a high degree of robot autonomy, the latter facilitates decision making and task allocation.
\item
\textit{Task representation level}. 
The second contribution is an integrated hierarchical representation of HRC processes employing First Order Logic (FOL) and AND/OR graphs to model static and dynamic aspects of HRC-related tasks. 
\end{itemize}

% \textcolor{blue}{-- Paper structure:}
%\textcolor{blue}{
%Previously in \cite{Darvish2018Mechatronics}, we proposed a flexible architecture for HRC scenarios, based on standard AND/OR graph and a simplified task manager to enable the team-level flexibility according to the human operator decisions which are recognised by wearable perceptual information. 
%In \cite{Darvish2018Roman}, we continued the precedent work by extending the previous HRC architecture and we have interleaved an online    robot simulator, to enhance the robot decision making capabilities. 
%To the best of authors' knowledge, for the first time, this paper proposes a First Order Logic (FOL) and hierarchical AND/OR graph, appropriate for the scalable representation of the complex assembly scenarios, and in specific for HRC scenarios.
%Moreover, we elaborate on the decision making framework of the robot founded on previous work to show how \textsc{FlexHRC+} enables the task and team level flexibilities.
%Finally, the proposed approaches are accompanied by extensive comprehensive demonstrations.
%}

With respect to our previous work \cite{Darvish2018Mechatronics, Darvish2018Roman}, this paper provides two significant improvements. 
The former is the use of a FOL-based encoding of information stored in a hierarchical AND/OR graph.
The FOL-based task representation has two important consequences.
First, it allows for modelling (part of) cooperation tasks on a non-grounded, terminological level, which can be therefore specified as a set of assertions anchored to objects in the robot workspace.
This allows for a more compact representation, with a consequent increased flexibility of the whole cooperation process.
Second, it decouples the cooperation task from the involved objects.
%, i.e., the same cooperation task and structure can be used with different objects, if compatible.
The latter foresees the use of hierarchical AND/OR graphs, which enable a great deal of modularity (when coupled with the FOL-based representation, graphs can be reused in different phases of the cooperation), as well as scalability.
%, which is essential to model and perform more complex scenarios, in terms of nodes, their connections, and eventually traversability. 

%Indeed, in \cite{Darvish2018Mechatronics} the architecture was based on standard AND/OR graph and a simplified task manager to enable the team-level flexibility according to the human operator decisions which are recognised by wearable perceptual information. Successively, in \cite{Darvish2018Roman}, we extended the architecture with the addition of  an online robot simulator, to enhance the robot decision making capabilities. However, this paper presents major improvements in the architecture, since First Order Logic (FOL) and hierarchical AND/OR graph are developed and integrated within the architecture. With these modifications, a scalable representation of the complex assembly scenarios is now possible, and in particular for HRC scenarios. Moreover, this paper elaborates on the decision making framework, showing how \textsc{FlexHRC+} enables task and team level flexibility. Finally, the paper presents extensive and comprehensive demonstrations supporting the new improvements.

The paper is organized as follows.
Section \ref{sec:RelatedWork} describes relevant state-of-the-art approaches in HRC processes. 
Section \ref{sec:Architecture} introduces the main traits of \textsc{FlexHRC+}.
We describe the FOL-based and hierarchical AND/OR graph task representation structure in Section \ref{sec:Cooperation Representation}, and the task management process in Section \ref{sec:Planner}.
In Section \ref{sec:Evaluation}, we describe the experimental scenarios and discuss relevant results.
Conclusions follow.

\section{Related Work}
\label{sec:RelatedWork}
% \textbf{[1 page]}

\textit{Task representation}.
Different approaches have been proposed to model HRC processes. 
Some of them are aimed at introducing aspects of social interaction \cite{Shah2011, Alami2017, Crandall2018}.
Others, similarly to the work described in this paper, focus explicitly on HRC processes for collaborative manipulation or assembly.
In this regard, the work in \cite{Johannsmeier2017} proposed a three-layer framework at the team, agent, and skill execution levels using AND/OR graphs. 
Task allocation is done by minimizing a cost function offline, whereas the reactive behaviour is managed at the control level. 
Such a work is extended in \cite{lamon2019capability} where the task allocation is modulated according to the co-worker ergonomics, and where the use of an augmented reality technology combined with the recognition of gestures of the human co-worker allows for an intuitive human-robot interaction.
In both contributions, the flexibility aspects and task allocation are restricted to the offline phase.
The uncertainties associated with the robot perception and the outcomes of robot actions are merely simplified at the control level. 
Therefore, the robot does not proactively make decisions. 
Moreover, the scalability of the solution cannot be easily determined.
In our work, we overcome these limitations by addressing the requirements $R_2$, $R_4$ and $R_5$ with an online task allocation for the human operator or the robot according to both human decisions as they unfold at run-time, and to the robot online simulation results. 
An example of these advancements is provided in Section \ref{sec:TableAssembly} and Section \ref{sec:IKEA}.

% In a similar work, \cite{Hawkins2014} proposed a probabilistic approach to anticipate the human actions in an assembly scenario using AND/OR graph task formalization. Therefore, the robot can adapt to the human online decisions, however it considers neither the intrinsic uncertainties endowed with its actions (partially fulfilling $R_2$) nor the variations in assembly scenario (missing $R_4$).

The work presented in \cite{Levine2014, Darvish2018Mechatronics} is aimed at recognising action sequences performed by human operators online, and to provide robots with methods to adapt accordingly. 
The recognition of such action sequences assumes actions to be completed \textit{before} they can be properly recognised. 
This is expected to introduce possibly unacceptable delays in the cooperation process, and therefore it might jeopardise efficiency and naturalness.
A slightly different approach, pursued in \cite{Hawkins2014}, employs probabilistic methods and AND/OR graphs to \textit{predict} human operator actions, thereby trading-off recognition performance and prediction accuracy.
In case of wrong predictions, the effectiveness of the overall cooperation process can be negatively affected.
Such an approach neither considers the intrinsic uncertainties of robot actions (partially fulfilling $R_2$) nor the variations in assembly scenarios (therefore missing $R_4$), while the table assembly scenario illustrated in Section \ref{sec:TableAssembly} meets those requirements.
%Rather than providing human operators with a high degree of freedom in how to carry out a given cooperative task, 
The approach proposed in \cite{Nikolaidis2017} envisions a \textit{dyadic}, mutual adaptation between human operators and robots.
Robots act as leaders guiding human operators towards an efficient task execution strategy.
It is no mystery that such an approach can lead to a lack of naturalness in the cooperation process.
While human action prediction, recognition, and adaptation are forms of implicit human-robot \textit{communication}, an explicit, speech-based communication is adopted instead in \cite{Alami2017, Shah2011}.
Aspects related to naturalness are enforced using speech-based communication only in principle.
In fact, this is done at the detriment of effectiveness and efficiency, since speech recognition can yield to dramatically poor results in industrial scenarios.

The work in \cite{Toussaint2016} implemented a concurrent, cooperative assembly task with relational Markov Decision Processes (MDPs), taking actions' duration into account.
%, which enables learning the cooperation processes as well. 
Using a probabilistic modelling of state transitions, the model can recover from failures in a reactive fashion. 
However, this is possible by disregarding perception uncertainties.
%associated with perception.
MDPs have been used in \cite{Claes2015, Crandall2018} to enforce adaptation to human operator behaviours online. 
Such an approach leads to purely reactive behaviours, which are considered indeed natural, but neither effective nor efficient.
In our case, 
%as pointed out while discussing the table assembly experiment, 
failures are avoided both with proactive decision trees and reactive traversal of an AND/OR graph. 
Finally, \cite{Toussaint2016} models the cooperation with a tree-like structure whereas, in our case, the AND/OR graph and its hierarchical structure makes the representation compact and modular.
This enables the implementation of complex scenarios, such as the one described in Section \ref{sec:IKEA}.

Task networks and approaches based on classical planning  \cite{cashmore2015, Capitanelli2018} are characterised by a natural description layer based on FOL \cite{Alami2017}.
Such a layer is expected to enforce effectiveness since it constitutes a close-to-human language used to associate semantics to each robot's action \cite{Russell2010}.
A cognitive approach based on an attention-based mechanism is proposed by \cite{Caccavale2016}, in which plans are generated using hierarchical task networks, and an attention-based system executes and monitors multiple plans while resolving possible conflicts. 
% In other to 
To demonstrate capabilities of the approach, a pick and place task in a simulated environment is shown, therefore eliminating intrinsic uncertainties characterising perception and action in real-world environments. 
Differently from our proposition put forth in Section \ref{sec:TableAssembly}, the experiments do not support reactive behaviour or proactive decision making at the team and task levels.
Similarly, \cite{paxton2017costar} proposed a collaborative system with a graphical user interface design toolkit for task automation and recognition based on first-order behaviour trees ($R_4$). 
%Differently from the table assembly scenario presented in Section \ref{sec:TableAssembly}, 
Such a collaborative system does not adapt to the intrinsic variability of human decisions or to the workspace status.
To accommodate for human preferences, an approach based on a scheduling and control framework has been introduced in \cite{wilcox2012}. 
In particular, an optimal scheduling policy embedding temporal constraints and human preferences are learned offline and executed online (partially meeting $R_2$). 
However, it does not support an online adaptation to the possible workspace variability (missing therefore $R_4$). 
Our contribution aims at overcoming these limitations with an online team and task level flexibility.
Moreover, none of these works provide evidence for scalability, as shown in Section \ref{sec:IKEA}.

In order to model action planning in HRC scenarios, hierarchical approaches have been used \cite{Alami2017}.
%As mentioned above, a hierarchical representation allows for the modelling of complex cooperation tasks efficiently, and it enforces modularity and scalability in the representation. 
The work in \cite{Tsarouchi2017} took an industry-oriented perspective for human-robot workplace design. 
%  the collaboration
It allows for scaling collaboration to complex scenarios using a three-layer task representation approach ($R_5$). 
Yet, differently from our approach, as demonstrated in Section \ref{sec:IKEA}, the one in \cite{Tsarouchi2017} limits scalability to three-layers at most.

\textit{Action planning}.
%\textcolor{red}{
%In order to model action planning in HRC scenarios, hierarchical approaches have been used in \cite{Hayes2016, Alami2017}.
As mentioned above, a hierarchical representation allows for the modelling of complex cooperation tasks efficiently, and it enforces modularity and scalability. 
%}
A few approaches consider the interplay between efficiency and naturalness in the cooperation process \cite{Alami2017, Caccavale2017, Sebastiani2017}. 
%\textcolor{red}{
%TNs and approaches based on classical planning are characterised by a natural description layer based on FOL \cite{cashmore2015, Alami2017, Capitanelli2018}, which is expected to enforce effectiveness since it constitutes a close-to-human language used to associate semantics to each robot action \cite{Russell2010}.
%}
%\sout{
%However, traversal and planning algorithms cannot guarantee explainable nor predictable robot action sequences, unless severe constraints are posed, for instance \textit{trajectory constraints} when planning algorithms are used.
%}
%\sout{
%FOL-based MDPs \cite{Boutilier2001, Yoon2002}, also referred to as \textit{relational} MDPs, are a tentative solution to trade-off these aspects, and indeed have been adopted to model cooperative assembly tasks in \cite{Toussaint2016}. 
%In a similar perspective, AND/OR graphs yield explainable results, and are amenable to be represented using FOL for the purpose of cooperative task representation \cite{Luger2009book}. 
%}

One of the main challenges to address in HRC scenarios is deciding how to allocate actions, either to the human operator, the robot, or in principle to both \cite{Goodrich2007}. 
%Action allocation is a necessary modelling choice relating task representation to task and motion planning combined together, and obviously it has great impact on effectiveness and efficiency.
When a human operator is given the freedom of autonomously deciding how to accomplish a task, action allocation cannot be defined beforehand and must be resolved online \cite{Miyata2002, Chen2014}. %Arai2002,
%In order to schedule resources and to allocate tasks to human operators or robots,
A multi-objective optimisation problem is typically formulated \cite{Chen2014, Tsarouchi2017}. 
It defines
%In particular, the allocation problem is addressed from the perspective of 
a \textit{utility} measure considering the \textit{quality} of an action result, its \textit{cost}, its associated \textit{cognitive load}, and the \textit{resources} needed for its completion \cite{Gerkey2004}.
Such optimisation problem is then resolved online for dynamic task allocation \cite{Shah2009, Chen2014, Giele2015}. % , wilcox2012
Other approaches are limited to offline solutions \cite{Tsarouchi2017, Johannsmeier2017}.
As described in \cite{Darvish2018Roman}, which is expanded in this paper, resolving action allocation online can be done only if the relevant parameters of the employed utility measure are either estimated beforehand or it is safe to assume they can be quantified while the cooperation process unfolds. 
%For example, in the approach described in \cite{Tsarouchi2017}, expected action completion times are estimated offline, whereas robot workspace reachability is assumed to be a function of the Euclidean distance between current and goal robot's end-effector poses.
%The approach discussed in \cite{Darvish2018Roman} proposes a fast in-the-loop simulation to estimate relevant utility parameters online, whereas in \cite{muller2007} it is suggested that simulations should be done \textit{a priori}.

It is noteworthy that the integration of task representation, online task planning, task allocation, and motion planning is expected to enhance the robustness to failures and the overall HRC process efficiency \cite{Darvish2018Roman}. 
It has been shown in \cite{Darvish2018Roman} that in-the-loop robot motion predictions are beneficial to a natural interaction, as opposed to the prediction of human operator motions \cite{Koppula2013, Mainprice2015}. 

\section{The \textsc{FlexHRC+} Architecture}
\label{sec:Architecture}
% \textbf{[1 page]}

\begin{figure}[!t]
\centering
\includegraphics[width=0.95\columnwidth]{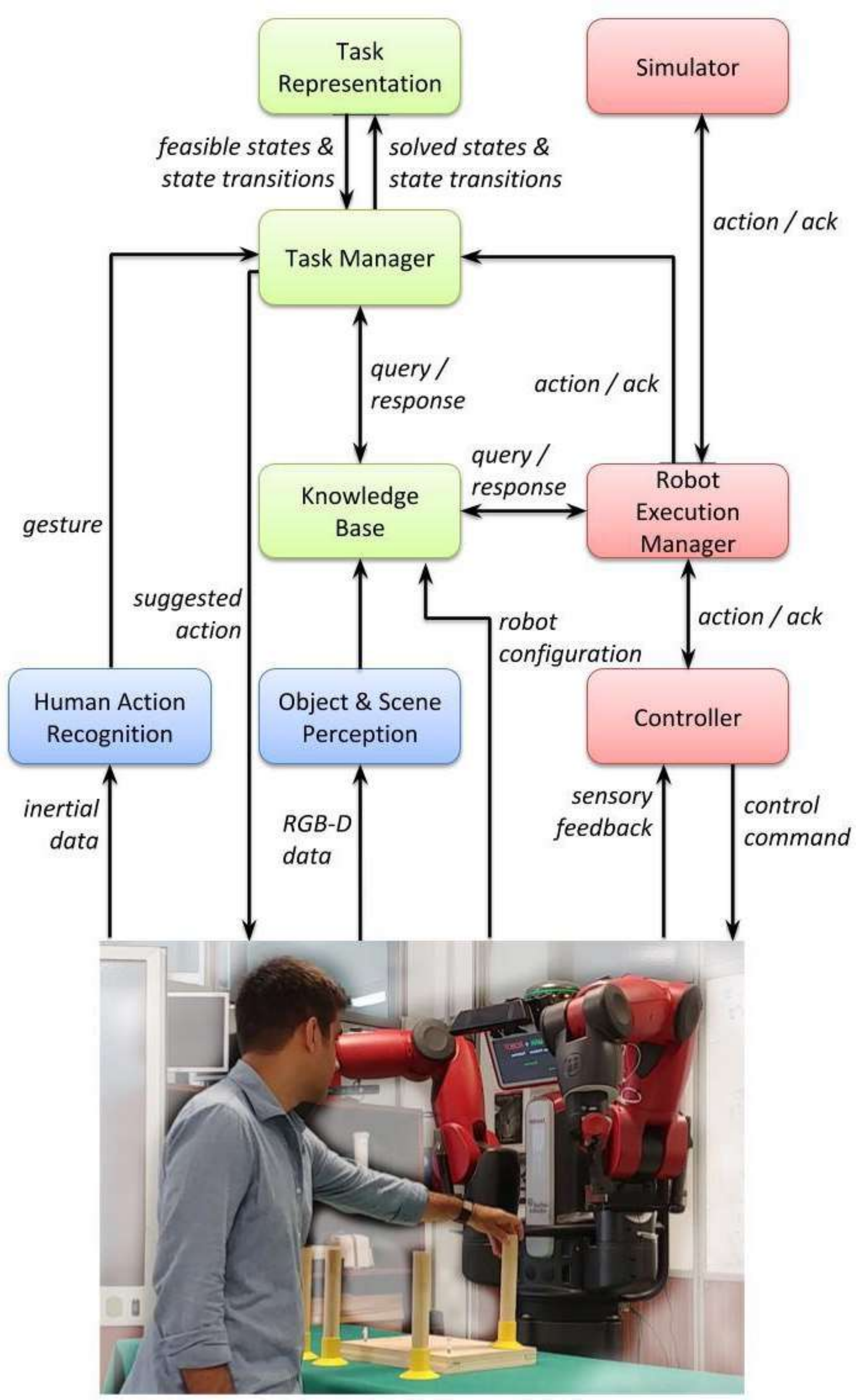}
\caption{\textsc{FlexHRC+}'s architecture: in \textit{green} the representation level, in \textit{blue} the perception level, in \textit{red} the action level.}
\label{fig:architecture}
\end{figure}

% \textcolor{blue}{-- one paragraph: describe the architecture of the HRC: } 

\textit{Overview}.
The architecture of \textsc{FlexHRC+} is organised in three levels, namely the representation level (depicted in \textit{green} in Figure \ref{fig:architecture}), the perception level (in \textit{blue}), and the action level (in \textit{red}).
The representation level maintains all the relevant information related to cooperative tasks via the \textit{Task Representation} module, and to the shared workspace in the \textit{Knowledge Base} module. 
The representation level orchestrates action planning and decision making for task execution, as well as action allocation, via the \textit{Task Manager} module. 
The perception level acquires information about the workspace in terms of objects and other entities therein using the \textit{Object and Scene Perception} module.
It is responsible for detecting and classifying actions performed by human operators as done in the \textit{Human Action Recognition} module. 
Starting from raw data, the perception level updates the representation level with relevant information about objects and human operator actions. 
Then, the action level serialises the execution of robot actions (via the \textit{Robot Execution Manager} module), performs in-the-loop robot action simulations in the \textit{Simulator} module, and controls online all robot motions using the \textit{Controller} module.

% unidirectional graphical interface, or robot display.
\textit{Representation level}.
The \textit{Task Representation} module maintains knowledge about all possible states and state transitions modelling cooperative tasks. 
The module also defines how the HRC process can progress by providing suggestions to the human operator or commands to the robot about what to do next. 
As anticipated above, we apply AND/OR graphs, primarily introduced by \cite{Mello1990}, to represent cooperative tasks \cite{ Darvish2018Mechatronics}, as better described in the next Section.
The module receives as input the current cooperation status from the \textit{Task Manager}, and therefore provides it with next action suggestions.
The \textit{Task Manager} module maps cooperation states as represented in the AND/OR graph structure to either human or robot actions.
%It provides the actions to the cooperative individuals.
The module plans for suggested states receiving appropriate information from the \textit{Task Representation} module, it grounds action parameters to actual values, and it performs action assignments on the basis of incoming perceptual information \cite{Darvish2018Mechatronics, Darvish2018Roman}.
The \textit{Knowledge Base} module explicitly represents the cooperation state and workspace-related perceptual information using custom data structures \cite{Darvish2018Roman}.

\textit{Perception level}.
The \textit{Human Action Recognition} module provides \textsc{FlexHRC+} with information about actions performed by human operators. 
The module models action templates using Gaussian Mixture Models and Gaussian Mixture Regression.
Models originate from a dataset of inertial data obtained using operator-worn sensors, and applies online 
%statistical distance metrics to determine the likelihood of the occurrence of an action \cite{Bruno2013}.
a pattern matching algorithm to detect and recognise meaningful actions as performed by human operators \cite{Bruno2013}.
To do so, the module receives an inertial data stream, and informs the \textit{Task Manager} about the detected and recognised human operator actions \cite{Bruno2014, Darvish2018Mechatronics}.
The \textit{Object and Scene Perception} module provides information about objects in the robot workspace and models them using a set of primitive shapes characterised by their geometrical characteristics. 
The module simply applies Euclidean distance to cluster a point cloud originating from an RGB-D sensor located on the robot body, it applies the Random Sample Consensus (RANSAC) algorithm to model those clusters as primitive shapes, and it determines their relevant features.
Additionally, Principal Component Analysis is used to find complementary object features for manipulation purposes.
Recognised objects and their features are maintained in the \textit{Knowledge Base} module \cite{Fischler1981, Wold1987, Buoncompagni15}.

\textit{Action level}.
The \textit{Robot Execution Manager} maps action commands issued by the representation level to \emph{control actions} fed to the \textit{Simulator} or the \textit{Controller} module. 
Each control action is associated with a hierarchy of \textit{equality} or \textit{inequality} control objectives.
These include reaching a desired end-effector position, avoiding obstacles and joint limits, or respecting the kinematic constraints imposed by a rigid object manipulated by two arms. 

In order to execute each control action, the \textit{Controller} module exploits a kinematic task-priority based framework \cite{Simetti2016Novel, Simetti2018}, which solves a sequence of prioritized optimization problems, computing the reference velocities for robot's actuators. 
The workspace-related feedback necessary to execute an action is received from the \textit{Knowledge Base}. 
The task-priority framework allows for the activation and deactivation of such inequality objectives as, e.g., maintaining a minimum distance from obstacles or from humans without over-constraining the solution when it is not necessary, hence those objectives can be put at the highest priority, increasing the safety of the system. 
Other safety and compliance features such as detecting and responding to contacts and impacts, preventing from applying continuous or excessive pressure, or keeping impact forces below design limits, are more related to the robot dynamic control, and are usually implemented within low-level control architectures.
Therefore, they are not the focus of this paper.

The in-the-loop \textit{Simulator} module replicates \textit{Controller} operations, integrating its output to simulate the robot behaviour online. 
It receives reference information from \textit{Robot Execution Manager}, and provides it with the results of the simulation, e.g., failure/success of a given command, action execution time, final robot pose, or estimated energy consumption \cite{Darvish2018Roman}.

\section{The Human-Robot Cooperation Model}
\label{sec:Cooperation Representation}
% [2-3 pages]

% \textcolor{blue}{-- Formalize the AND/OR graph problem Description:} 

\subsection{Single-layer AND/OR Graphs}
\label{sec:single_layer}
In order to formalise the human-robot cooperation process in \textsc{FlexHRC+}, we adopt AND/OR graphs. 
An AND/OR graph allows for an easy decomposition of problems and procedures in their building blocks (as parts of the graph), as well as the logic relationships among them (i.e., the  graph connectivity).
Since in AND/OR graphs the root node represents the solution to the problem being modelled, solving it means traversing the graph from leaf nodes to the root node according to its structure.
AND/OR graphs can take limited forms of non-determinism or uncertainty into account \cite{Mello1990, Luger2009book, Russell2010} via the availability of various branches leading to the solution. 
In previous work \cite{Darvish2018Mechatronics, Darvish2018Roman}, such representation has been adopted to model the online behaviour of simple cooperation processes.
In this paper we systematise the original formulation and extend it along two directions:
first, we provide a conceptualisation of AND/OR graphs compatible with a FOL-based task representation framework, which allows us to better model cooperation \textit{templates}; 
second, we introduce and analyse the benefits of hierarchical AND/OR graphs to support modular, scalable, and flexible task representation.

An AND/OR graph $G$ can be formally defined as a tuple $\langle N, H \rangle$ where $N$ is a set of $|N|$ nodes, and $H$ is a set of $|H|$ hyper-arcs. 
An hyper-arc $h \in H$ induces two sets of nodes, namely the set $N_c(h) \subset N$ of its \textit{child} nodes, and the singleton $N_p(h) \subset N$ made up of a \textit{parent} node, such that
\begin{equation}
h: N_c(h) \rightarrow N_p(h).
\label{eq:hyper_arc_trans_simple}
\end{equation}
For the scenarios we consider, at the semantic level each node $n \in N$ represents a particular \textit{state} related to the cooperation between a human operator and a collaborative robot.
Each hyper-arc $h \in H$ represents a (possibly) many-to-one \textit{transition} among states, i.e., activities performed by human operators or robots that make the cooperation move forward.
In \textsc{FlexHRC+}, a node $n \in N$ is associated with a conjunction $S(n)$ of literals, such that
\begin{equation}
S(n) = s_1 \land \ldots \land s_k \land \ldots \land s_{|S(n)|},
\label{eq:state_definition}
\end{equation}
where each literal $s_k$ may consist of variables, constants or logic predicates, also negated.
As it will be described later, each literal can be considered a representation \textit{fragment} related to the cooperation state defined by $n$.
As a consequence, we will refer to $S(n)$ as the \textit{cooperation state} represented in $n$. 
It is noteworthy that a given $S$ does not have to include only grounded literals, i.e., constants or grounded predicates.
It can include variables as well, and as such the corresponding node can be treated as a \textit{class} of states (or cooperation templates) at the \textit{Task Representation} level \cite{Russell2010}.

Using the definition of \textit{state} in \eqref{eq:state_definition}, it is possible to better specify the state transition in \eqref{eq:hyper_arc_trans_simple}.
This is a relationship induced on
% the 
hyper-arc $h$ between a set
%$R(h)$
of requirements made up by joining all the literals defining states $S(n_k)$ associated with all the nodes $n_k \in N_c(h)$, and a set
%$E(h)$
of effects made up by the state $S(n)$ associated with
% the 
single node $n \in N_p(h)$, such that
\begin{equation}
h: S(n_1) \land \ldots \land S(n_k) \land \ldots \land S(n_{|N_c|}) \rightarrow S(n).
\label{eq:hyper_arc_trans_complex}
\end{equation}

%FOL AND/OR graph not only increases the computational efficiency of the task representation but also makes the design of it easier for a user.
%As an example, if there are a number of identical objects in the workspace to manipulate, the FOL representation allows lower level modules (Task Manager)  taking care of making the decision for an object manipulation and grounding the truth values.

The relation among child nodes in hyper-arcs is the logical \textit{and}, whereas the relation between different hyper-arcs inducing on the same parent node is the logical \textit{or}.
Different hyper-arcs inducing on the same parent node represent alternative ways for a cooperation process to move on.
We define $n \in N$ as a \textit{leaf} node if $n$ is not acting as a parent node for any hyper-arc, i.e., if $h \in H$ does not exist such that $n \in N_p(h)$.
Similarly, we define $n$ as a \textit{root} node if it is the only node that is not a child node for any hyper-arc, i.e., if $h \in H$ does not exist such that $n \in N_c(h)$. 

Each hyper-arc $h \in H$ implements the transition in \eqref{eq:hyper_arc_trans_complex} by checking the truth values associated with all the requirements defined by nodes in $N_c(h)$, executing a number of actions associated with $h$, and generating effects compatible with the cooperation state of the parent node.
In particular, each hyper-arc $h \in H$ is responsible for executing an ordered set $A(h)$ of \textit{actions}, such that
\begin{equation}
A(h) = (a_1, \ldots, a_k, \ldots, a_{|A|}; \preceq),
\end{equation}
where the precedence operator $\preceq$ defines the pairwise expected order of action execution.
Before an hyper-arc $h$ is executed, all actions $a \in A(h)$ are marked as \textit{undone}, and we refer to this using a predicate ${\textsf{done}(a) \leftarrow false}$.
When one action $a$ is executed either by the human operator or the robot, its status changes to \textit{done} as ${\textsf{done}(a) \leftarrow true}$.
An hyper-arc $h \in H$ is marked as \textit{solved}, i.e., ${\textsf{solved}(h) \leftarrow true}$ \textit{iff} all actions $a \in A(h)$ are done in the expected order.

In a similar way, nodes $n \in N$ may be associated with a (possibly ordered) set of \textit{processes} $P(n)$, i.e., 
\begin{equation}
P(n) = (p_1, \ldots, p_k, \ldots, p_{|P|}; \preceq).
\end{equation}
Differently from actions, which are instrumental to perform transitions from one cooperation state to another, and must be necessarily executed by human operators or robots, processes are robot behaviours which do not imply any state transition.
They are used to control physical or other non-functional variations of some quantity over time. 
An example may be a robot behaviour aimed at keeping a certain object in a given pose or configuration using two grippers, the effects of external forces notwithstanding, because the related cooperation state assume that object pose.
Processes are meant at being executed by robots, whereas a human operator may carry them out occasionally.
Each process is characterised by a priority, or precedence.
%If the robot and a human operator are managing a process, then they are occupied, and therefore they cannot carry out another action or process in the meantime.
Actions and processes are associated with hyper-arcs and states, respectively. 
However, there is no causal relationship between actions and processes, hence, they can be executed in any order, according to the priority or preference.
When a node $n \in N$ is reached, all of its processes are \textit{activated}, i.e., ${\textsf{activated}(p) \leftarrow true}$ for each $p \in P(n)$. 
A process $p \in P(n)$ can be \textit{deactivated} when certain process-specific termination conditions are met, i.e., ${\textsf{activated}(p) \leftarrow false}$. 
A node $n$ is marked as \textit{met}, i.e., ${\textsf{met}(n) \leftarrow true}$, if all the associated processes are deactivated if necessary in the prescribed order, or $P(n)$ is an empty set.

Using these definitions, it is possible to introduce the notion of feasibility for nodes and hyper-arcs.
A node $n \in N$ is \textit{feasible}, which we refer to as $\textsf{feasible}(n) \leftarrow true$, \textit{iff} a solved hyper-arc $h \in H$ exists, for which $n \in N_p(h)$, and $\textsf{met}(n) \leftarrow false$, i.e.,  
\begin{equation}
\exists h \in H. \left(\textsf{solved}(h) \cap n \in N_p(h) \cap \neg \textsf{met}(n)\right).
\label{eq:feasible_node}
\end{equation}
All leaf nodes in an AND/OR graph are usually feasible at the beginning of the human-robot cooperation process, which means that the cooperation itself can be performed in many ways, and is not constrained to follow certain sequences of operations.
In a similar way, an hyper-arc $h \in H$ is \textit{feasible}, i.e., $\textsf{feasible}(h) \leftarrow true$, \textit{iff} for each node $n \in N_c(h)$, $\textsf{met}(n) \leftarrow true$ and $\textsf{solved}(h) \leftarrow false$, i.e.,
\begin{equation}
\forall n \in N_c(h).\left(\textsf{met}(n) \cap \neg \textsf{solved}(h)\right).
\label{eq:feasible_hyper_arc}
\end{equation}
Once an hyper-arc $h_i \in H$ is solved, all other feasible hyper-arcs $h_j \in H\setminus\{h_i\}$, which share with $h_i$ at least one child node, i.e., $N_c(h_i) \cap N_c(h_j) \neq \emptyset$, are marked as unfeasible, in order to prevent the cooperation process to consider alternative ways to cooperation that have become irrelevant.

The human-robot cooperation process is modelled as a \textit{graph traversal} procedure.
Starting from a set of leaf nodes, it must reach the root node by selecting hyper-arcs and reaching states in one of the available sequences, depending on the feasibility statuses of nodes and hyper-arcs.
To this aim, each node $n \in N$ is associated with a \textit{weight} $w(n)$, and each hyper-arc $h \in H$ is similarly associated with a weight $w(h)$.
Weights are related to the number, difficulty or time-to-completion of actions/processes, and to other more qualitative metrics related to human operator preferences \cite{Buoncompagni2018Roman}.
Nodes or hyper-arcs with lower weights are privileged compared to others with higher weights. 
Weights are identified through several demonstrations of expert users.
Then, a \textit{cooperation path} $cp$ induced by $G$ is a set of nodes and hyper-arcs, such that 
\begin{equation}
cp = (n_1, \ldots, n_k, h_1, \ldots, h_l), 
\label{eq:cp}
\end{equation}
which represents a particular way to connect leaf nodes to the root node.
We refer to the set of cooperation paths induced by $G$ as $CP(G)$, where each element $cp \in CP$ is in the form described by \eqref{eq:cp}.
According to the structure of the modelled human-robot cooperation task, multiple cooperation paths may exist, meaning that multiple ways to solve the task may be equally legitimate. 
Each cooperation path $cp \in CP$  can be associated with an overall cost $c(cp)$, such that
\begin{equation}
c(cp) = \sum_{j=1}^{k} w(n_j) + \sum_{j=1}^{l} w(h_j).
\label{eq:pathCost}
\end{equation}
The different cooperation paths in $CP$ can be ranked according to their overall costs.
Two cooperation paths $cp_i$ and $cp_j \in CP$ are \textit{equal} \textit{iff} the corresponding sets of nodes and hyper-arcs are the same, and are \textit{equivalent} \textit{iff} their corresponding overall costs are the same. 

The traversal procedure dynamically follows the cooperation path that at any time is characterised by the lowest cost. 
The traversal procedure suggests to human operators actions in the hyper-arcs that are part of the path, and sends to robots actions they must execute.
Human operators can override the suggestions at any time, executing different actions, which may cause the system to be in a cooperation state not part of the current cooperation path.
When this situation is detected, \textsc{FlexHRC+} tries to progress from that state onwards \cite{Darvish2018Mechatronics,Darvish2018Roman}.
This mechanism enables \textsc{FlexHRC+} to pursue an optimal path leading to the solution, while it allows human operators to choose alternative paths when they deem it fit.
As long as the human-robot cooperation process unfolds, and the AND/OR graph is traversed, we refer with $N_f$ and $H_f$ to the sets of \textit{currently} feasible nodes and hyper-arcs, respectively.
In fact, the actual elements of these two sets depend on the
% particular
evolution of the cooperation process.

We say that an AND/OR graph $G$ is \textit{solved}, and we write $\textsf{solved}(G) \leftarrow true$, \textit{iff} its root node $r \in N$ is met, i.e., $\textsf{met}(r) \leftarrow true$.
Otherwise, if the condition $N_f \cup H_f = \emptyset$ (i.e., there are no feasible nodes nor hyper-arcs) then the human-robot cooperation process is failed, because there is no feasible cooperation path leading to the root.
It is noteworthy that representations based on AND/OR graphs, when updated online, do not require the full knowledge of the robot workspace.
%, nor actions that are irrelevant for the current cooperation path. 
In fact, while a given cooperation path is followed, the traversal algorithm only needs knowledge about feasible nodes and hyper-arcs for making the task progress.

% \textcolor{blue}{Solved, unsolved states or state-transition, actions done/undone, feasible/infeasible node, hyper-arc: }
%Following this section, first we introduce the new algorithms of the FOL AND/OR graph; subsequently, we introduce the Hierarchical AND/OR graph.

%\subsection{Single-layer AND/OR Graph Traversal Procedure}

% \textcolor{blue}{ -- Two phases of normal And/or graph : offline and online: - (I) offline phase is similar, (II) updated  online phase algorithms for optimal AND/OR graph }

%% offlinePhase()
% \begin{algorithm}[!t]
% \begin{algorithmic}[1]
% \caption{$\textsc{offlinePhase()}$}
% \label{alg:OfflinePhase}
% \Require {An AND/OR graph $G = \langle N, H \rangle$}
% \Ensure {A data structure encoding $G$, a set $\Phi=\{\langle x, c(x) \rangle\}$ of suggestions}
% \State $G \leftarrow \textsc{loadDescription()}$
% \State $\textsf{solved}(G)$ $\leftarrow$ \textit{false}
% \State $\textsc{updateGraphFeasibility(G)}$
% \State $CP \leftarrow \textsc{generateAllPaths(G)}$
% \State $\Phi \gets \textsc{findSuggestions(G)}$
% \end{algorithmic}
% \end{algorithm}

The AND/OR graph structure presented in this paper is based on the one introduced in \cite{Darvish2018Mechatronics}.
Notable differences are the possibility of allowing for multiple hyper-arcs connecting the same child nodes to a parent node, ensuring the minimum cost returned from each hyper-arc or node, and supporting a FOL-based task representation. 
The first feature allows the AND/OR graph to model different state transitions from one cooperation state to another, 
the second one ensures an optimal, predictable, and therefore explainable robot behaviour, whereas
the last one increases the overall expressive power of the representation structure. 

The single-layer AND/OR graph traversal procedure is composed of two phases, the first being offline and the second online. 
% The offline phase is described in Algorithm \ref{alg:OfflinePhase}. 
% The functions \textsc{loadDescription()} and \textsc{generateAllPaths()} in Algorithm \ref{alg:OfflinePhase} are described with great detail in \cite{Darvish2018Mechatronics}.
%, while the new version differs since it computes the path cost using \eqref{eq:pathCost}.
% In summary, \textsc{loadDescription()} generates the data structure $G$ (line 1), 
% the graph $G$ is setup as unsolved (line 2), 
% all feasibility statuses for nodes and hyper-arcs are checked (line 3),
% the set $CP$ of cooperation paths is generated (line 4), 
% and suggestions for next actions (in terms of nodes and hyper-arcs) are computed on the basis of path costs defined as in \eqref{eq:pathCost} in line 5.
The offline phase loads the description of the AND/OR graph, generates the data structure $G$, sets the graph status as unsolved.
Later, the feasibility of all nodes and hyper-arcs is checked, the set $CP$ of cooperation paths is generated, and suggestions for next actions (in terms of nodes and hyper-arcs) are computed as defined in \eqref{eq:pathCost}.

%% updateGraphFeasibility()
% \begin{algorithm}[!t]
% \begin{algorithmic}[1]
% \caption{$\textsc{updateGraphFeasibility()}$}
% \label{alg:updateGraphFeasibility}
% \Require {An AND/OR graph $G = \langle N,H \rangle$}
% \Ensure {The feasibility sets $N_f$ and $H_f$}
% \State $N_f = \emptyset$
% \State $H_f = \emptyset$
% \ForAll{$n \in N$}
% \State $(N_f, H_f) \leftarrow$ \textsc{updateNodeFeasibility($n$, $N_f$, $H_f$)}
% %\If{$\textsf{feasible}(n)$ = \textit{true}}
% %\State $N_f \leftarrow N_f \cup n$
% %\EndIf
% \EndFor
% \ForAll{$h \in H$}
% \State $(N_f, H_f) \leftarrow$ \textsc{updateHyperarcFeasibilty($h$, $N_f$, $H_f$)}
% %\If{$\textsf{feasible}(h)$ = \textit{true}}
% %\State $H_f \leftarrow H_f \cup h$
% %\EndIf
% \EndFor
% %\State \textbf{return} ($N_f$, $H_f$)
% \end{algorithmic}
% \end{algorithm}

%% UpdateFeasibilityNode()
\begin{algorithm}[!t]
\begin{algorithmic}[1]
\caption{\textsc{updateNodeFeasibility()}}
\label{alg:UpdateFeasibilityNode}
\Require {A node $n$, feasibility sets $N_f$ and $H_f$}
\Ensure {Updated feasibility sets $N_f$ and $H_f$}
\State $\textsf{feasible}(n) \leftarrow false$
\State $N_f \leftarrow N_f \backslash \{n\}$
\If{$\textsf{met}(n)$}
\ForAll{$h$ s.t. $n \in N_c(h)$}
\If{$\textsf{solved}(h)$}
\State $\textsf{feasible}(h) \leftarrow false$
\State $H_f \leftarrow H_f \backslash \{h\}$
\Else
\State $\textsf{feasible}(h) \leftarrow true$
\State $H_f \leftarrow H_f \cup \{h\}$
\ForAll{$n'$ s.t. $n' \in N_c(h)$}
\If{$\neg \textsf{met}(n')$}
\State $\textsf{feasible}(h) \leftarrow false$
\State $H_f \leftarrow H_f \backslash \{h\}$
\State \textbf{break}
\EndIf
\EndFor
\EndIf
\EndFor
\Else
\If{$N_c(n) = \emptyset$}
\State $\textsf{feasible}(n) \leftarrow true$
\State $N_f \leftarrow N_f \cup \{n\}$
\Else
\ForAll{$h$ s.t. $n \in N_p(h)$}
\If{$\textsf{solved}(h)$}
\State $\textsf{feasible}(n) \leftarrow true$
\State $N_f \leftarrow N_f \cup \{n\}$
\State \textbf{break}
\EndIf
\EndFor
\EndIf
\EndIf
\State \textbf{return} ($N_f$, $H_f$)
\end{algorithmic}
\end{algorithm}

%% UpdateFeasibiltyHyperarc()
\begin{algorithm}[!t]
\begin{algorithmic}[1]
\caption{\textsc{updateHyperarcFeasibility()}}
\label{alg:UpdateFeasibilityHyperarc}
\Require {A hyper-arc $h$, feasibility sets $N_f$ and $H_f$}
\Ensure {Updated feasibility sets $N_f$ and $H_f$}
\State $\textsf{feasible}(h) \leftarrow$ \textit{false}
\If{$\textsf{solved}(h)$}
\State $n \leftarrow N_p(h)$
\If{$\neg\textsf{met}(n)$}
\State $\textsf{feasible}(n) \leftarrow$ \textit{true}
\State $N_f \leftarrow N_f \cup \{n\}$
\EndIf
\ForAll{$n$ s.t. $n \in N_c(h)$}
\ForAll{$h'$ s.t. $n \in N_c(h')$}
\State $\textsf{feasible}(h') \leftarrow$ \textit{false}
\State $H_f \leftarrow H_f \backslash \{h'\}$
\EndFor
\EndFor
\Else
\State $\textsf{feasible}(h) \leftarrow$ \textit{true}
\State $H_f \leftarrow H_f \cup \{h\}$
\ForAll{$n$ s.t. $n \in N_c(h)$ }
\If{$\neg\textsf{met}(n)$}%\textit{or} $f(n)=$\textit{false}}
\State $\textsf{feasible}(h) \leftarrow$ \textit{false}
\State $H_f \leftarrow H_f \backslash \{h\}$
\State \textbf{break}
\ElsIf{$\exists h'$ s.t. $\textsf{solved}(h') \wedge n \in N_c(h')$}
\State $\textsf{feasible}(h') \leftarrow$ \textit{false}
\State $H_f \leftarrow H_f \backslash \{h'\}$
\EndIf
\EndFor
\EndIf
\State \textbf{return} ($N_f$, $H_f$)
%\State \textbf{return} ($N_f$, $H_f$)
%\State \textit{return}
\end{algorithmic}
\end{algorithm}

The update of feasibility statuses of all the involved nodes and hyper-arcs, i.e., populating the corresponding sets $N_f$ and $H_f$, is done simply by iteratively invoking two functions, namely \textsc{updateNodeFeasibility()} and \textsc{updateHyperarcFeasibility()}.
% It is worth discussing the behaviour of function \textsc{updateGraphFeasibility(G)}, described by Algorithm \ref{alg:updateGraphFeasibility}, which updates the feasibility statuses of all involved nodes and hyper-arcs, and populates the corresponding sets $N_f$ and $H_f$.
% The Algorithm works simply by iteratively invoking two functions, namely \textsc{updateNodeFeasibility()} (line 4) and \textsc{updateHyperarcFeasibilty()} (line 7).
The two functions are further developed in Algorithm \ref{alg:UpdateFeasibilityNode} and Algorithm \ref{alg:UpdateFeasibilityHyperarc}. 
%If, after these two functions are called, a given node or hyper-arc is feasible, it is included in $N_f$ or $H_f$, respectively.
Given a node $n$ and a hyper-arc $h$ the two Algorithms use such predicative knowledge on $n$ and $h$ as the values of $\textsf{feasible}(n)$, $\textsf{feasible}(h)$, $\textsf{met}(n)$, and $\textsf{solved}(h)$ to update the feasibility of graph nodes and hyper-arcs, respectively, therefore producing updated sets $N_f$ and $H_f$.
In Algorithm \ref{alg:UpdateFeasibilityNode}, lines 3-19 update the feasibility status of a relevant hyper-arc $h$ when $n \in N_c(h)$ and $\textsf{met}(n)$ holds true.
In case node $\textsf{met}(n)$ holds false (lines 20-32), lines 21-23 change the node feasibility when it does not have any child node, i.e., if $n$ is not a parent of any hyper-arc.
Lines 24-31 check for a solved hyper-arc connected to node $n$, and in case at least one of such hyper-arcs exists, then it is marked as feasible. 
In Algorithm \ref{alg:UpdateFeasibilityHyperarc}, lines 2-13 update the feasibility statuses when the hyper-arc $h$ is solved. 
The feasibility of the $h$'s parent node (line 3) is updated in lines 4-7. 
The feasibility of all the hyper-arcs having a common set of child nodes with $h$ is updated in lines 8-13. 
Lines 14-27 check the feasibility of the unsolved hyper-arc $h$.
If a child node of $h$ (line 17) is not met (lines 18-21) or there is another solved hyper-arc $h'$ with a set of child nodes common with $h$ (lines 22-25), the hyper-arc $h$ becomes infeasible. 
Finally, \textsc{findSuggestions()} in Algorithm \ref{alg:findSuggestions} determines the set $\Phi$ of suggestions whose elements are generically indicated using $x$, and their associated cost $c(x)$.
There might be different paths from $N_f$ or $H_f$ to the root of $G$.
Therefore, the AND/OR graph is expected to provide the \textit{minimum cost} among all these cooperation paths to ensure optimality.
The cost $c(x)$ for a node or hyper-arc is the minimum cost of the cooperation path $cp$ which the node or the hyper-arc belongs to.
Therefore the Algorithm guarantees the optimality of the AND/OR graph because for all nodes or hyper-arcs in $N_f$ or $H_f$, respectively, it holds that:
\begin{equation}
\label{eq:andor_optimality}
c(x) = \min_{x \in cp}{c(cp)}.
\end{equation}
In Algorithm \ref{alg:findSuggestions}, lines 3-11 return feasible nodes and the minimum cost of the cooperation paths which include them.
The same applies to hyper-arcs in lines 12-20.

%% findSuggestions()
\begin{algorithm}[!t]
\begin{algorithmic}[1]
\caption{\textsc{findSuggestions()}}
\label{alg:findSuggestions}
\Require {An AND/OR graph $G = \langle N,H \rangle$}
\Ensure {A set $\Phi = \{\langle x, c(x) \rangle\}$ of suggestions}
\State $\Phi = \emptyset$
\State $cost \leftarrow 0$
\ForAll{$n \in N$ s.t. $\textsf{feasible}(n)$}
\State $cost \leftarrow \inf$
\ForAll{$cp \in CP(G)$ s.t. $n \in cp$}
\If{$cost < c(cp)$}
\State $cost \leftarrow c(cp)$
\EndIf
\EndFor
\State $\Phi \leftarrow  \Phi \cup \{\langle n, cost \rangle\}$
\EndFor
\ForAll{$h \in H$ s.t. $\textsf{feasible}(h)$}
\State $cost \leftarrow \inf$
\ForAll{$cp \in CP(G)$ s.t. $h \in cp$}
\If{$cost < c(cp)$}
\State $cost \leftarrow c(cp)$
\EndIf
\EndFor
\State $\Phi \leftarrow  \Phi \cup \{\langle h, cost \rangle\}$
\EndFor
\State \textbf{return} $\Phi$
\end{algorithmic}
\end{algorithm}

%% OnlinePhase()
\begin{algorithm}[!t]
\begin{algorithmic}[1]
\caption{\textsc{onlinePhase()}}
\label{alg:OnlinePhase}
\Require {An AND/OR graph $G = \langle N, H \rangle$, feasibility sets $N_f$ and $H_f$, the met set $N_m$, the solved set $H_s$}
\Ensure {An updated AND/OR graph $G$, updated feasibility sets $N_f$ and $H_f$, A set $\Phi = \{\langle x, c(x) \rangle\}$ of suggestions}
%\If{\textit{query} = $true$}
\State $n_r \leftarrow$ \textsc{getRoot($G$)}
\ForAll{$n \in N_m$}
\State \textsc{metNode($n$, $G$)}
\State \textsc{updateNodeFeasibility($n$, $N_f$, $H_f$)}
\EndFor
\ForAll{$h \in H_s$}
\State \textsc{solvedHyperarc($h$, $G$)}
\State \textsc{updateHyperarcFeasibility($h$, $N_f$, $H_f$)}
\EndFor
\If{$\textsf{met}(n_r)$}
\State $\textsf{solved}(G) \leftarrow$ \textit{true}
\State \textbf{return}
\EndIf
\State \textsc{updateAllPaths($G$, $N_m$, $H_s$)}
\State $\Phi \leftarrow$ \textsc{findSuggestions($G$)}
\State \textbf{return}
%\EndIf
%\State \textit{return}
\end{algorithmic}
\end{algorithm}
% During online execution, when a node is met or a hyper-arc is solved, the \textit{Task Manager} module queries the AND/OR graphs to get updated $N_f$ and $H_f$, as well as the associated costs, in order to make the cooperation progress.
% This is done by Algorithm \ref{alg:OnlinePhase}. 
The online phase follows Algorithm \ref{alg:OnlinePhase}.
The two sets of met nodes and solved hyper-arcs are referred to as $N_m$ and $H_s$, respectively.
Upon the reception of the \textit{Task Manager}'s query, the Algorithm updates node and hyper-arc statuses (in terms of \textsf{solved}, \textsf{met} and \textsf{feasible} predicates) in lines 2-9. 
The \textsf{solved} status for the whole AND/OR graph is checked.
If the root node is met, then the graph is marked as solved (line 11).
Otherwise, line 14 updates all the path weights, which include nodes in $N_m$ and hyper-arcs in $H_s$. 
In line 15 the new feasible nodes and hyper-arcs, and their associated costs, are made available.
In the Algorithm, the functions \textsc{metNode($n$, $G$)} and \textsc{solvedHyperarc($h$, $G$)} check first if $\textsf{feasible}(n)$ or $\textsf{feasible}(h)$ hold true, then update $G$ by $\textsf{met}(n) \leftarrow true$ and $\textsf{solved}(h) \leftarrow true$.
In particular,
% Algorithm \ref{alg:updateAllPaths}
\textsc{updateAllPaths($G$, $N_m$, $H_s$)}
updates the cooperation path costs at each query.
For a given cooperation path $cp \in CP$, the path cost $c(cp)$ at each moment is the cost of traversing it from the current to the root state. 
Initially, all the path costs are computed from the leaves to the root using \eqref{eq:pathCost}.
When a node or hyper-arc belonging to a given cooperation path is met or solved, its overall cost is reduced of an amount related to its weight, i.e.,
% (lines 3 and 8).
\begin{equation}
\forall x \in N_m , H_s : c(cp) = c(cp) - w(x). 
\label{eq:graph_transition}
\end{equation}

% %% updateAllPaths
% \begin{algorithm}[!t]
% \begin{algorithmic}[1]
% \caption{\textsc{updateAllPaths()}}
% \label{alg:updateAllPaths}
% \Require {An AND/OR graph $G = \langle N, H \rangle$, the met set $N_m$, the solved set $H_s$}
% \Ensure {An updated set $CP$}
% \ForAll{$n \in N_m$}
% \ForAll{$cp \in CP$ s.t. $n \in cp$}
% \State $c(cp) = c(cp) - w(n)$
% \EndFor
% \EndFor
% \ForAll{$h \in H_s$}
% \ForAll{$cp \in CP$ s.t. $h \in cp$}
% \State $c(cp) = c(cp) - w(h)$
% \EndFor
% \EndFor
% %\State \textit{return}
% \end{algorithmic}
% \end{algorithm}

\subsection{Hierarchical AND/OR Graphs}
\label{sec:hierarchical}
% \textcolor{blue}{ -- Hierarchical AND/OR graph}

% %\textcolor{blue}{Add the description of the AND/OR graph updates wrt previous version:}

% % \textcolor{blue}{How give the weight of a node/hyper arch when it is asked:}

% \textcolor{blue}{Add the algorithm of hierarchical and/or graph:}
% The hierarchical representation and planning of the cooperation is coming from both representation and planning level. In the \textit{Task Representation} there are a set of AND/OR graphs $\{G_1, ..., G_M\}$, which the \textit{Planner}  can call each of these graphs with unique name while progressing the cooperation.
% The higher level graph represents a transition from the initial states to final state in generic terms, while the lower level graphs details the state transition.

The use of hierarchical AND/OR graphs in the context of HRC tasks has two motivations. 
The first is related to the computational complexity of single-layer AND/OR graphs, while the second is related to flexibility and scalability requirements.
It has been shown that AND/OR graphs are characterised by a polynomial time complexity in the number of nodes and hyper-arcs \cite{Laber2008}.
The problem of determining whether a solution in terms of a path from the set $N_L$ of leaf nodes to the root node is NP-hard \cite{Sahni1974}. 
In the online phase of HRC tasks, being able to quickly determine and select an alternative cooperation path to take into account human operator preferences is therefore of the utmost importance. 
On the computational side, this means reducing the number of nodes and hyper-arcs which the \textit{Task Manager} module must reason upon. 
Different real-world operations are structured as mandatory or alternative sets of human or robot actions, which can be seen as \textit{atomic}.
Being able to identify and re-use the same sub-sequences of operations in different parts of the same HRC process or as part of different processes is expected to enhance flexibility, because such sub-sequences can be easily substituted if needed, and scalability, since the overall complexity can be increased maintaining a manageable representation overhead.
%Laber, E. S., A randomized competitive algorithm for evaluating priced and/or trees, Theorical Computer Science, v. 401, n. 1, p. 120-130, 2008.
% Sahni S., Computationally related problems, SIAM Journal on Computing, v. 3, n. 4, p. 262-279, 1974.

Analogously to single-layer AND/OR graphs, a hierarchical AND/OR graph $\mathcal{H}$ is defined as a tuple $\langle \Gamma, \Theta \rangle$ where $\Gamma$ is an ordered set of $|\Gamma|$ AND/OR graphs, such that:
\begin{equation}
\Gamma = \left(G_1, \ldots, G_{|\Gamma|}; \preceq \right),
\label{eq:gamma_set}
\end{equation}
and $\Theta$ is a set of $|\Theta|$ transitions between couples of AND/OR graphs.
In \eqref{eq:gamma_set}, the AND/OR graphs are pairwise ordered according to their \textit{depth} level.
With a slight abuse of notation, we associate a depth level $l$ to an AND/OR graph $G$ and we indicate it with $G^l$, the highest level being $l=0$.   
AND/OR graphs with increasing depth levels are characterised by a decreasing level of abstraction, i.e., deeper graphs model HRC more accurately.
Transitions in $\Theta$ define how different AND/OR graphs in $\Gamma$ are connected, and in particular model the relationship between any $G^l$ and a deeper connected graph $G^{l+1}$. 

It is necessary to better define transitions.
If we recall \eqref{eq:hyper_arc_trans_complex} and we contextualise for an AND/OR $G^l = \langle N^l, H^l \rangle$, we observe that a given hyper-arc in $H^l$ represents a mapping between the set of its child nodes and the singleton parent node.  
We can think of a generalised version of such a mapping to encompass a whole AND/OR graph $G^{l+1} = \langle N^{l+1}, H^{l+1} \rangle$, where the set of child nodes is constituted by the set $N^{l+1}_L$ of leaf nodes, and the singleton parent node by the graph's root node $r^{l+1} \in N^{l+1}$, such as: 
\begin{equation}
G^{l+1}: S(n^{l+1}_1) \land \ldots \land S(n^{l+1}_k) \land \ldots \land S(n^{l+1}_{|N^{l+1}_{L}|}) \rightarrow S(r^{l+1}).
\label{eq:graph_trans_complex}
\end{equation}
A transition $T$ can defined between a hyper-arc $h^l \in H^l$ and an entire deeper AND/OR graph $G^{l+1}$, such that:
\begin{equation}
T: h^l \rightarrow G^{l+1}, 
\label{eq:graph_transition}
\end{equation}
subject to the fact that appropriate mappings can be defined between the set of child nodes of $h^l$ and the set of leaf nodes of the deeper graph, i.e.,
\begin{equation}
M_1: N_c(h^l) \rightarrow N_L \in N^{l+1}, 
\label{eq:mapping_1}
\end{equation}
and the singleton set of parent nodes of $h^l$ and the root node of the deeper graph, i.e.,
\begin{equation}
M_2: N_p(h^l) \rightarrow r^{l+1} \in N^{l+1}. 
\label{eq:mapping_1}
\end{equation}
Mappings $M_1$ and $M_2$ must be such that the conjunction of literals of nodes in $N_c(h^l)$ and the conjunction of literals of leaves in $G^{l+1}$ is \textit{semantically equivalent}.
They should be the same or representing the same information with a different depth of representation, for example each literal of nodes in $N_c(h^l)$ may correspond to one or more literals of nodes in $N^{l+1}_L$. 
The same applies for the root of $G^{l+1}$ and $N_p(h^l)$.
Once these mappings are defined, it easy to see that $\mathcal{H}$ has a tree-like structure, where graphs in $\Gamma$ are nodes and transitions in $\Theta$ are edges.

%To represent it, we use the notation $g_{l+1}=\textsf{nestedGraph}(h)$, where $g_{l+1}$ is the \textit{nested} graph at level $l+1$ and $h \in g_{l}$ is a hyper-arc belonging to the higher level graph $g$ at level $l$.

% In the hierarchical AND/OR graph, a hyper-arc $h$ at level $l$ may contain another AND/OR graph at level $l+1$.

%The hierarchical AND/OR representation should agree with \textit{continuity} principle in the physical world; therefore:
%\begin{equation}
%\begin{aligned}
%&{g_{l+1}: S(n_1) \land \ldots \land S(n_k) \land \ldots \land S(n_{|N_{Leaf|}}) \iff}\\ &{ h \in g_{l}: S(n_1) \land \ldots \land S(n_k) \land \ldots \land S(n_{|N_c|})}
% \rightarrow S(n).
% \rightarrow S(r).
%\label{eq:graph_trans_complex}
%\end{aligned}
%\end{equation}
%and $g_{l+1}: S(r) \iff h \in g_{l}: S(n_p)$. This means that the conjunction of literals of the child nodes of the $h$ and the conjunction of the literals of leaves of the graph $g_{l+1}$ should be equivalent. The same applies for the root of the graph $g_{l+1}$ and parent node of the hyper-arc $h$.

An AND/OR graph $G^l$ is feasible, and we refer to it as $\textsf{feasible}(G^l)$ if it has at least one feasible node or hyper-arc.
If a transition $T$ exists in the form \eqref{eq:graph_transition}, a hyper-arc $h^l \in H^l$ is feasible \textit{iff} the associated deeper AND/OR graph $G^{l+1}$, is feasible:
\begin{equation}
\forall T.\left(\textsf{feasible}(G^{l+1}) \leftrightarrow \textsf{feasible}(h^l)\right).
\label{eq:feasibility_hierarchical_graph}
\end{equation}
Shen the hyper-arc $h^l$ becomes feasible in $G^l$, the nodes in $N^{l+1}_L$ of $G^{l+1}$ become feasible as well.
Furthermore, the hyper-arc $h^l$ is solved \textit{iff} the associated deeper AND/OR graph $G^{l+1}$ is solved:
\begin{equation}
\forall T.\left(\textsf{solved}(G^{l+1}) \leftrightarrow \textsf{solved}(h^l)\right).
\label{eq:solve_hierarchical_graph}
\end{equation}

For all hyper-arcs in $H^{l}$ for which a transition $T$ towards $G^{l+1}$ exists, we must define how to compute the related weight.
If we define $cp^{l+1,*}$ the cooperation path in $G^{l+1}$ characterised by the lowest cost, we easily define:
\begin{equation}
w(h^l) = c\left(cp^{l+1,*}\right).
\label{eq:hierarchical_andor_optimality}
\end{equation}
In this case the weight is attributed using an optimistic strategy, because as per change of the cooperation path in $G^{l+1}$ it may happen that the actual $w(h^l)$ is underestimated.

%Moreover, the hierarchical AND/OR graph formalism avoids the curse of dimensionality, enhances the computational time, and enables the scalable architecture.

%As an example, let us consider a manufacturer assembles a smartphone using the hierarchical AND/OR graph. In the highest level AND/OR graph, the user may define main parts of the cell phone such as a display, battery, system-on-a-chip (SoC), memory, modem, camera, and sensors, and in which order they are assembled. Using the hierarchical AND/OR graph, the user defines in the lower level graph the details of the different processes to assemble a part to another one. The assembling details of each part (which is composed of other detailed sub-parts) can be nested and defined in other lower level AND/OR graphs.

Similarly to the single-layer case, hierarchical AND/OR graphs are used in two phases, first offline and then online.
A transition $T$ is modelled using a function in the form $G^{l+1} = \textsc{lowerGraph}(h^l)$, whereas the inverse relationship is obtained using $h^l = \textsc{upperHyperarc}(G^{l+1})$.
The offline phase
%, similar to what already described for the single-layer case in Algorithm \ref{alg:OfflinePhase}, 
first loads the description of the highest-level AND/OR graph $G^0$.
Considering any nesting level $l$, if a hyper-arc $h \in H^l$ is associated with a deeper AND/OR graph description $G^{l+1}$ by a transition, the Algorithm calls the function $\textit{OfflinePhase}()$ on $G^{l+1}$ to build it before going on with $G^{l}$. 
%When $g_{l+1}$ is constructed, it continues the construction of the higher-level graph $g_{l}$.
%This procedure applies to all the layers of the hierarchical AND/OR graph.
% If a hyper-arc of $g_1$ nests another lower-level graph $g_2$, the same procedure applies.

Algorithm \ref{alg:OnlinePhaseHierarchical} describes the workflow associated with the hierarchical AND/OR graph during online execution.
%When a query arrives in the hierarchical AND/OR graph, nodes and hyper-arcs should contain the trace to the corresponding graph name they belong, $n_i \in g_i$ and $h_j \in g_j$.
%A simple basic AND/OR graph may nest in different places of hierarchical AND/OR graph; therefore with the trace of the graph, we can find the graph object in which $h_j$ or $n_i$ belongs.
%As a result when a query arrives, the met nodes or solved hyper-arc should pair with their corresponding graph trace as shown in requirements of Algorithm \ref{alg:OnlinePhaseHierarchical}.
Whenever the status of the HRC process needs updating, the graph representation is updated starting from all sets $N_{i,m}$ of met nodes, and the sets $H_{i,s}$ of solved hyper-arcs, for all AND/OR graphs in $\Gamma$ (lines 3-12).
After node statuses are updated in lines 5-6, the Algorithm checks whether any graph is solved (line 7).
If this holds true and the solved graph is not the root graph of $\mathcal{H}$, then the associated higher-level hyper-arc is labelled as solved (line 9) and then included in the corresponding set of solved hyper-arcs $H_{i,s}$.
Lines 14-19 update the feasibility statues for all solved hyper-arcs.
Then, if the root node of the root graph is met, then the whole graph is solved (line 21) and the Algorithm terminates (line 22).
Otherwise, all cooperation paths are updated (line 25), and the set $\Phi^H$ of next suggestions is found (line 30), as better described in Algorithm \ref{alg:findSuggestionHierarchical}. 
It is noteworthy that $\Phi^H$ includes $\Phi$ and adds to each triplet the graph label containing the node or hyper-arc.  

%Beside the mentioned differences, Algorithms \ref{alg:OnlinePhase} and \ref{alg:OnlinePhaseHierarchical} are similar.

%% onlinePhaseHierarchical Algorithm
\begin{algorithm}[!t]
\begin{algorithmic}[1]
\caption{\textsc{onlineHierarchicalPhase()}}
\label{alg:OnlinePhaseHierarchical}
\Require {A hierarchical AND/OR graph $\mathcal{H} = \langle \Gamma, \Theta \rangle$, feasibility sets $N_{i,f}$ and $H_{i,f}$, the met set $N_{i,m}$, the solved set $H_{i,s}$ for each $G_i \in \Gamma$}
\Ensure {An updated hierarchical AND/OR graph $\mathcal{H}$, updated feasibility sets $N_{i,f}$ and $H_{i,f}$ for each $G_i \in \Gamma$, a set $\Phi^H = \{ \langle x, c(x), g(x)\}$ of suggestions}
\State $G^0 \leftarrow$ \textsc{getRootGraph($\Gamma$)}
\State $r^0 \leftarrow$ \textsc{getRoot($G^0$)}
\ForAll{$G_i \in \Gamma$}
\ForAll{$n \in N_{i,m}$}
\State \textsc{metNode($n$, $G_i$)}
\State \textsc{updateNodeFeasibility($n$, $G_i$)}
\If{$\textsf{solved}(G_i)$ and $G_i \neq G^0$}
\State $h \leftarrow$ \textsc{upperHyperarc($G_i$)}
\State $\textsf{solved}(h) \leftarrow$ \textit{true}
\State $H_{i,s}$ $\leftarrow$ $H_{i,s} \cup \{h\}$
\EndIf
\EndFor
\EndFor
\ForAll{$G_i \in \Gamma$}
\ForAll{$h \in H_{i,s}$}
\State \textsc{solvedHyperarc($h$, $G_i$)}
\State \textsc{updateHyperarcFeasibility($h$, $N_{i,f}$, $H_{i,f}$)}
\EndFor
\EndFor
\If{$\textsf{met}(r^0)$}
\State $\textsf{solved}(G^0) \leftarrow$ \textit{true}
\State \textbf{return}
\EndIf
\ForAll{$G_i \in \Gamma$}
\State \textsc{updateAllPaths($G_i$, $N_{i,m}$, $H_{i,s}$)}
\EndFor
\State $N_f \leftarrow N_{1,f} \cup \ldots \cup N_{|\Gamma|,f}$
\State $H_f \leftarrow H_{1,f} \cup \ldots \cup H_{|\Gamma|,f}$
%\ForAll{$(n,g) \in N_m$}
%\State MetNode($n$, $g$)
%\State UpdateFeasibilityNode($n$, $g$)
%\If{$solved(g)=true$ and $g \neq G$}
%\State $(h',g')$ $\leftarrow$ HigherLevelGraph($g$)
%\State $H_s$ $\leftarrow$ $H_s \cup (h',g')$
%\EndIf
%\EndFor
%\ForAll{$(h, g) \in H_s$}
%\State SolvedHyperarc($h$, $g$)
%\State UpdateFeasibiltyHyperarc($h$, $g$)
%\EndFor
%\If{$\textsf{met}(n_r)$ s.t. $n_r \in G$}
%\State $solved(G) \leftarrow$ \textit{true}
%\State \textit{return}
%\EndIf
%\State updateAllPaths($\mathcal{CP}$,$N_m$, $H_s$ )
\State $\Phi^H = \emptyset$
\State $\{\langle x, c(x), g(x)\rangle\} \leftarrow$ \textsc{findSuggestions($\mathcal{H}$, $\Phi^H$)}
\State \textbf{return}

%x \in N_f$ \textit{or} $ H_f \} \leftarrow$ findSuggestionHierarchical($G$)
%\State \textit{return}
%\State \textit{return}
\end{algorithmic}
\end{algorithm}
%%%%%%%%%%%%%%%%%%%%%%%%%%%%%%%%%%%%%%%%%%%%

% %%%%%%%%%%%%%%%%%%% updateAllPaths %%%%%%%%%%%%%%%%%%%%%%%%%
% \begin{algorithm}[!t]
% \begin{algorithmic}[1]
% \caption{updateAllPathsHierarchical()}
% \label{alg:updateAllPathsHierarchical}
% \Require {An AND/OR graph $G$, set of solved nodes and associated graph pairs $N_s$ and solved hyper-arcs and associated graph pairs $H_s$}
% \Ensure {Updated path costs}

% \ForAll{$(n,g) \in N_s$ }
% \ForAll{$p \in P(g)$ such that $n \in p$ }
% \State	$c(p)= c(p)-c(n)$
% \EndFor
% \EndFor
% \ForAll{$h \in H_s$ }
% \ForAll{$p \in P(g)$ such that $h \in p$ }
% \State	$c(p)= c(p)-c(h)$
% \EndFor
% \EndFor

% \State \textit{return}
% \end{algorithmic}
% \end{algorithm}
% %%%%%%%%%%%%%%%%%%%%%%%%%%%%%%%%%%%%%%%%%%%%

Algorithm \ref{alg:findSuggestionHierarchical} finds first feasible nodes part of an optimal cooperation path (lines 2-12), as well as the associated cost and graph.
A similar operation is done in lines 13-32 for hyper-arcs.
In this case it is necessary to check whether a transition exists towards a deeper AND/OR graph.
If this is not the case, the hyper-arc is stored as a suggestion.
Otherwise, the associated graph is determined and the function is recursively called on it.
Finally, the minimum cost from the parent node of a hyper-arc to the root node of the corresponding graph is computed in line 27, and the suggestion updated accordingly.
Figure \ref{fig:kitchen_assembly} provides an example of a hierarchical AND/OR graph for a kitchen assembly scenario. 
On the left hand side, the first layer includes $9$ nodes, with the root node being on top, and $5$ hyper-arcs. 
Hyper-arcs $h_1$ and $h_2$ can be further specialised as second layer graphs, e.g., the one in the mid of the Figure. 
This is characterised by $41$ nodes and $14$ hyper-arcs (partly depicted in the Figure), two of which are further specialised as a third layer on the right.

% check
%% findHierarchicalSuggestions
\begin{algorithm}[t!]
\begin{algorithmic}[1]
\caption{\textsc{findSuggestions()}}
\label{alg:findSuggestionHierarchical}
\Require {A hierarchical AND/OR graph $\mathcal{H} = \langle \Gamma, \Theta \rangle$, a set $\Phi^H=\{\langle x, c(x), g(x) \rangle \}$ of suggestions}
\Ensure {An updated set $\Phi^H$}

\State \textit{cost} $\leftarrow$ 0

\ForAll{$G_i = \langle N_i, H_i \rangle \in \Gamma$}
\ForAll{$n \in N_i$ s.t. $\textsf{feasible}(n)$}
\State \textit{cost} $\leftarrow$ inf
\ForAll{$cp \in CP(G_i)$ s.t. $n \in cp$}
\If{\textit{cost} $< c(cp)$}
\State \textit{cost} $\leftarrow$ $c(cp)$
\EndIf
\EndFor
\State $\Phi^H \leftarrow \Phi^H \cup \{\langle n, cost, G_i\rangle\}$
\EndFor
\EndFor

\ForAll{$G_i = \langle N_i, H_i \rangle \in \Gamma$}
\ForAll{$h \in H_i$ s.t. $\textsf{feasible}(h)$}
\State \textit{cost} $\leftarrow$ inf
\ForAll{$cp \in CP(G_i)$ s.t. $h \in cp$}
\If{\textit{cost} $< c(cp)$}
\State \textit{cost} $\leftarrow$ $c(cp)$
\EndIf
\EndFor
\If{\textsc{lowerGraph($h$)} = null}
\State $\Phi^H \leftarrow \Phi^H \cup \{\langle h, cost, G_i\rangle\}$
\Else
\State $G_j \leftarrow$ \textsc{lowerGraph($h$)}
\State $\Phi^H \leftarrow$ \textsc{findSuggestion($G_j$, $\Phi^H$)}, with $x \in N_{j,f} \cup H_{j,f}$
\ForAll{$\langle x, c(x), g(x) \rangle \in \Phi$}
\State $cost \leftarrow cost - w(h) + c(x)$
\State $\Phi^H \leftarrow \Phi^H \cup \{ \langle x, cost, g(x) \rangle \}$
\EndFor
\EndIf
\EndFor
\EndFor
\State \textbf{return} $\Phi^H$
\end{algorithmic}
\end{algorithm}

\section{Reasoning on the Cooperation Model}
\label{sec:Planner} 
% [2-3 pages]
% \textcolor{blue}{ -- Describe the module generally, introduce the key ideas based on previous paper (state-action table, simulation or decision tree, action using pddl formalization): the key point of this module is arrive from states to the grounded actions}
% \textcolor{blue}{-- Formalize the Planner Module based on the description}
% The \textit{Planner} module maps the state space to the action space such that the cooperation performance maximizes. % and predict the robot behavior by simulation in order to
All reasoning tasks on the cooperation model are carried out within the \textit{Task Manager} module.
The module receives the sets of feasible states or hyper-arcs from the \textit{Task Representation} module, and determines the sequence of actions for each hyper-arc (or the sequence of processes for states), it grounds relevant parameters to actions, and allocates actions to human operators or robots to maximise a utility indicator.
%The module reactively adapts to the human online decisions.

% solving the feasible states/ state transitions (Later on for simplicity we call both state and state transition as state). 

\subsection{Reasoning upon First Order Logic based AND/OR Graphs}
\label{sec:PlannerFormalization} 

Differently from what happens in standard AND/OR graphs, \textsc{FlexHRC+} encodes \textit{actions} in hyper-arcs using a notation compliant to the Planning Domain Definition Language (PDDL) formalism \cite{cashmore2015}.
%The principal attribute of \textit{Task Manager} is the notion of an \textit{action}.
% We start formalization of the module by defining the action.
In \textsc{FlexHRC+}, an action $a$ contributes to a transition modelled as a hyper-arc \eqref{eq:hyper_arc_trans_complex}.
To do so, it acts on a set $param(a)$ of parameters to anchor, and it maps a set $pre(a)$ of precondition literals in conjunction (possibly defining cooperation states) to a set $eff(a)$ of effect literals, which are part of other cooperation states in the graph or are intermediate literals.
The set of effect literals can be split into two disjoint sets, namely a set of literals $eff^+(a)$ holding true after the action has been executed, and a set of literals $eff^-(a)$ not holding anymore after the action. 
%models is a transition applied on its parameters from a set of physical states (conjunction of literals), namely  \textit{preconditions} to the new states, namely \textit{effects}, $a: s\rightarrow s'$. 
Therefore, an action in \textsc{FlexHRC+} can be defined as:
\begin{equation}
a = \left(param(a), pre(a), eff^+(a), eff^-(a)\right).
\end{equation}
%where the sets of $Parameter$, \textit{PreCondition} $\subset s$, and \textit{Effects} $\subset s'$ are conjunction of literals in which their interpretations are known. 

\noindent Each action $a$ is associated with a set $agents(a)$ enumerating the agents (either human operators or robots), which may be responsible for performing $a$.
It is noteworthy that such a set may be a singleton (e.g., when only one agent can be allocated to the action), may define a set of possibilities (e.g., when a human operator or a robot may be tasked with the same action), or may define a list of agents which may be required to perform the action \textit{jointly}.
%If the number of human or robot, performing an action $a_i$, is bigger than one, we call it a \textit{joint action}, in which they should have a \textit{joint convention} to perform $a_i$.
Using such a formalisation, although the semantics associated with the literals is known, they may not be anchored to any real object in the robot workspace at the modelling level.
%We assume the grounding of the parameters does not affect the planning problem, otherwise, we solve a new planning problem.
%According to the definition provided in Section \ref{sec:Cooperation Representation}, a process $P_i \in P$ does not affect the states of the cooperation used in \textit{Task Representation}, therefore $p_i: s\rightarrow s$.
%Given the feasible states or state transitions coming from the \textit{Task Representation}, all the preconditions and effects are defined as well as the domain of the actions. Taking into account the interpretation of the variables in \textit{Task Manager}, we define a planning problem to find the ordered sequence of actions $(a_1, ...., a_n)$ from the preconditions to the effects such that the state transition is executed. In this paper, we assume the sequence of these actions is provided.

Using the sequences of the actions for all the feasible states and hyper-arcs with their associated costs, \textit{Task Manager} creates a data structure called \textit{Action-State table} \cite{Darvish2018Mechatronics}.
The table keeps the information of the chosen state to follow and the progress of the associated action executions.
%A thorough description of Action-State table is given in \cite{Darvish2018Mechatronics}.
% Using this table, the robot simply chooses the minimum cost cooperation path to follow. 
Given 
% \sout{ the set of feasible states or hyper-arcs, and the associated costs,}
the Action-State table,
the \textit{Task Manager} either proactively selects which hyper-arc to follow, or adapts to human preferences as soon as their actions are duly recognised. 
In both cases, in order to reach the goal node in the graph, it identifies a cooperation path to follow with the minimum cost according to \eqref{eq:pathCost}.
Once the \textit{Task Manager} selects a feasible hyper-arc that is part of the minimum cost path, to perform each action associated with the hyper-arc, \textsc{FlexHRC+} must anchor non-grounded literals in action definitions, and allocate agents to each action.
%It is the process of selecting the state or state transition with the minimum costs (optimal state), grounding the literals, assigning the actions to the human or robot, and the examination of the optimal state execution.
To do so, first updated information about the workspace is retrieved online from the \textit{Knowledge Base} module.
Then, all the possible literal groundings are determined, as well as all the possible combinations of agents, which may be tasked with the set of ordered actions associated with the selected hyper-arc.
%of the actions and the possible agents (human or robot) who can perform all the actions of the optimal state. 
Using such information, a decision tree is automatically generated \cite{Darvish2018Roman}, whose various branches are related to the diverse parameter groundings and agents who can perform the actions.
% Using such information, a decision tree is automatically generated \cite{Darvish2018Roman}, whose various branches are related to the diverse parameter groundings and agents the actions can take.
% and agents who can perform the actions.
%  agents \sout{an action} \textcolor{blue}{the  set of ordered actions} can take.
%ich is different from the classical decision tree introduced in \cite{Russell2010}.
All branches in the tree are ranked using a utility function, i.e., a metric to estimate performance and
% the
quality of actions execution \cite{Gerkey2004}.
% \sout{The value is generated by simulating each action multiple times varying the simulation parameters on the basis of the branches of the decision tree.}
The value is generated using the utility function for each leaf of the decision tree, by simulating all the ordered set of actions associated with the selected hyper-arc, varying action parameters or assigned agents.

%Then, we examine the execution of all the branches by simulating the robot actions online. 
%We use the results of the simulations to compute the \textit{utility value} of the decision tree branches,
%which is a metric to estimate the performance and the quality of the execution of a state transition \cite{Gerkey2004}. 

In general terms, a utility function to determine the best course of action for a cooperation model can be defined as the weighted sum of several robot-centred or human-centred metrics, e.g., the closest simulated distance to obstacles, the maximum joint accelerations, the maximum velocities, the overall execution time, which can be used to evaluate the overall execution \textit{quality}. 
In our work, we have opted to consider only the execution time. 
Hence, given a utility function $J$ and a branch $b$ of the decision tree, the utility function $J(b)$ is defined as to maximise the inverse of the total execution time associated with the branch $b$, namely:
\begin{equation}
\label{eq:utility}
J(b) = \epsilon(b) \times \frac{1}{\sum\limits_{i=1}^{K} t_{i}},
\end{equation}
% \epsilon(success, failure)
where $K$ is the number of actions of the selected hyper-arc or branch, $t_i$ is the execution time of the $i$-th action in the branch $b$, and the unit function $\epsilon(b)$ equals to one if all  actions in $b$ are executed successfully, to zero otherwise.
A simulation is successful if the agent can reach its given goal in a defined time interval.
We ground the parameters of the optimal state and assign the actions to an agent based on the branch with the maximum utility value.
If the utility value of all the branches in the decision tree becomes zero, the \textit{Task Manager} sets the optimal state as infeasible and attempts for a new optimal state. 
With this method, the \textit{Task Manager} proactively avoids cooperation from failing and increases the overall robustness of the architecture.
Action success or failure, and -- in case of success -- the time it takes to complete an action is determined by the \textit{Simulator} module, which mimics the robot kinematic behaviour using the currently available knowledge of the robot state and its workspace.
If disturbances are limited and the robot model is known with sufficient precision, the likelihood of having consistent results in simulation and with the real robot is increased. 
%Therefore, the estimates in Equation \ref{eq:utility} is realistic and similar to what will happen in reality. 
One may decide to simulate the robot dynamics, but in our experience this option, depending on the given task, may not provide more detailed information to compute the utility value because of uncertainties in the interaction with the environment.

\begin{figure*}[t]
\centering
\includegraphics[width=0.99\textwidth]{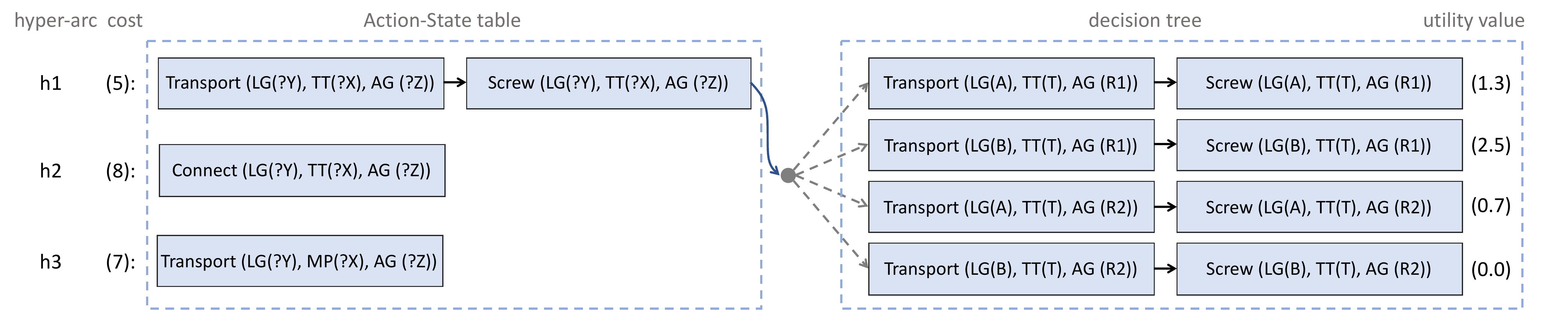}
\caption{An example of the interconnections among feasible states or hyper-arcs, the Action-State table, and the decision tree. LG stands for $\textsc{Leg}$, TT stands for $\textsc{Tabletop}$, MP stands for $\textsc{MiddlePose}$ (in front of the human operator), which are action parameters, whereas AG stands for the set of $\textsc{Agent}$s.}
\label{fig:TaskManagerExample}
\end{figure*}

As an example, let us consider a table assembly task where a tabletop $T$ must be connected to either leg $A$ or $B$, as shown in Figure \ref{fig:TaskManagerExample}.
The example will be further explored in Section \ref{sec:Evaluation}.
The task can be performed by a human operator $H$ or either robot $R_1$ or $R_2$, interchangeably.
When the \textit{Task Manager} receives the set of feasible hyper-arcs and their associated costs from the \textit{Task Representation} module, it generates the corresponding Action-State table.
Each row of the Action-State table shows the sequence of atomic actions to be carried out in the given state or hyper-arc.
In the example shown in the Figure, the cooperation cost of hyper-arc $h_1$ is the lowest (with a value of $5$) among the three feasible hyper-arcs $h_1$, $h_2$, and $h_3$. 
Therefore, \textit{Task Manager} selects $h_1$ as the optimal state, and generates the corresponding decision tree (on the right hand side of the Figure).
Using the information stored in the knowledge base, the \textit{predicate} $\textsc{Tabletop(?x)}$ can be grounded to $T$ only, and $\textsc{Leg(?y)}$ can be grounded to one among $A$ or $B$, whereas $\textsc{Agent(?z)}$ can be assigned to either $R_1$ or $R_2$.
In this case, the decision tree ramifies to four branches in total.
\textit{Task Manager} computes the utility value by simulating all the actions in each branch, and the one with maximum utility, i.e., the second branch with $J = 2.5$, is selected to ground the parameters (leg $B$ and tabletop $T$) and responsible agents (robot $R_1$) of the set of actions associated with the given hyper-arc. 
Grounding the literals online according to the perceived information of the workspace allows the architecture to adapt to task-level variations, therefore enforcing flexibility.
If, during an actual cooperation process, a human operator attempts to achieve a different feasible state, e.g., $h_2$, then the \textit{Task Manager} adapts to such a decision and grounds the predicates accordingly. 
Moreover, if the robot cannot perform a given action in a certain amount of time
despite a successful simulation, assuming $\textsc{Transport}\left(\textsc{Leg($B$)}, \textsc{Tabletop($T$)}, \textsc{Agents($R_1$)}\right)$ in $h_1$, then the \textit{Task Manager} stops the robot from executing its current task, sets the optimal hyper-arc to follow as infeasible, and finds a new optimal hyper-arc $h_3$ among the available ones, making the system more robust to failures and to environment uncertainties.
However, if the execution of all the hyper-arcs fails, the collaboration as a whole fails.

It is noteworthy that a discussion of formal properties could be relevant for (i) the traversal procedure associated with the AND/OR graph, and (ii) the decision tree. 
In the first case, formal properties have been demonstrated by Nguyen and Szalas \cite{NguyenSzalas2009} related to AND/OR graphs when used as \textit{context-free semi-Thue systems}, as in our case, including succinctness, correctness and completeness.
%The fact that we use a hierarchical version of AND/OR graphs does not modify such proof, because for each hierarchical graph it is possible to obtain a 1-layer, flatten (and redundant) graph an increased number of nodes and hyper-arcs. 
In the second case, the decision tree is generated to model different assignments to variables, and as such is akin to brute force, although encoded in a compact representation.

\subsection{Behaviour of the Task Manager}
\label{sec:PlannerAlgorithm} 

\begin{figure*}[!t]
\centering
\includegraphics[width=0.95\textwidth]{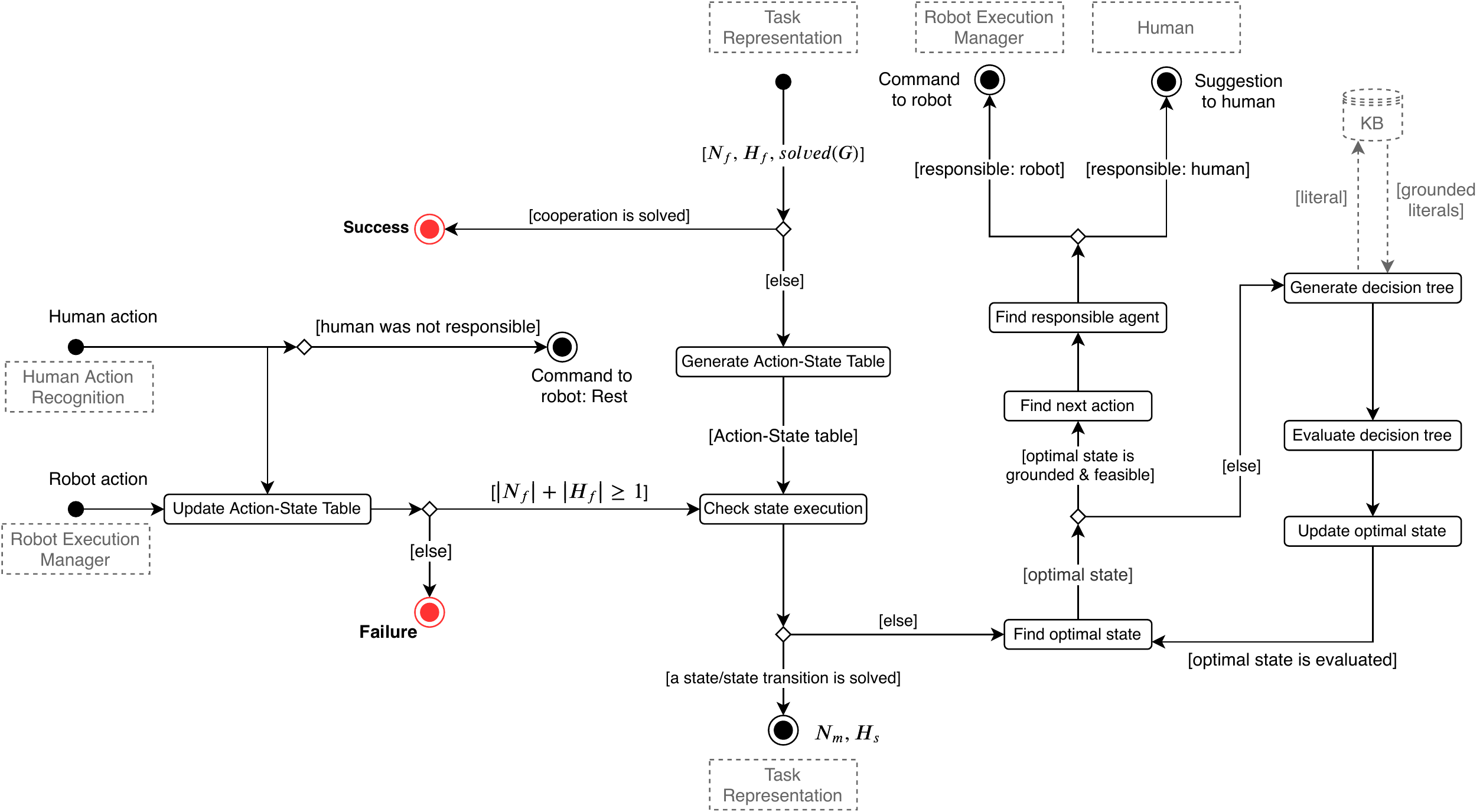}
\caption{A flowchart depicting the \textit{Task Manager} online phase.}
\label{fig:plannerModuleOnline}
\end{figure*}

The \textit{Task Manager} is organised in two phases, respectively offline and online. 
Offline, the \textit{Task Manager} initialises the list of agents participating in the collaboration scenario, all action descriptions, and the possible robots or humans who can perform each action.
An action may be executed by different agents individually, or jointly. 
However, action assignment is done online, and the necessary skill to perform each robot action is instructed in \textit{Robot Execution Manager} or the \textit{Controller}.
Then, the \textit{Task Manager} loads the set of action sequences for all states or hyper-arcs involved in the cooperation process.
Using such information, we create the necessary data structures for the online execution. 
Figure \ref{fig:plannerModuleOnline} shows a flowchart associated with the online phase.
The phase starts with an empty query from the \textit{Task Representation} module.
When the response to such a query is available, together with the set of feasible states or state transitions, the \textit{Task Manager} first checks whether the cooperation graph is successfully solved.
Otherwise, it generates the Action-State table as described above and checks for a met state or a solved state transition in the \textit{Check state execution} function.
Afterwards, among all feasible states, it finds the optimal state using the function \textit{Find optimal state}, and checks if the actions in such a state are grounded or assigned, as well as whether the robot can successfully execute the actions in the simulation.
In order to ground the optimal state action parameters and to assign actions to agents, the \textit{Task Manager} first generates the decision tree in \textit{Generate decision tree}, then it simulates all the actions associated with its branches using a breadth-first search algorithm, and then it computes the utility value for all the branches in \textit{Evaluate decision tree}.
Finally, using the function \textit{Update optimal state}, it checks for the maximum utility value, it grounds all action parameters, and assigns the actions to the agents in the optimal state.
If the maximum utility value is zero, the function \textit{Update optimal state} sets the current optimal state as infeasible. 
In this case, \textit{Find optimal state} selects another state with minimum cost among
% the
others available in the Action-State table, and again loops in-between simulation and decision tree evaluation.
Eventually,
% the
\textit{Find next action} finds the first action in the grounded or evaluated optimal state that is not done yet, and 
% the
\textit{Find responsible agent} is called to assign the command to an agent.
% Afterwards, among the feasible states, we find the optimal state by function \texttt{Find Optimal State()}. To ground the truth values of the actions of the optimal state, we generate the decision tree and find the utility value for all the branches. Function \texttt{Generate Simulation Tree()} generates a tree of the possible combinations for grounding the literals in the optimal state and task assignment to the cooperative individuals; \texttt{Simulate Next Action()} finds the first non-simulated action using Breadth-first search. When a given simulation is done, \texttt{Update Simulation Tree()} updates the decision tree.

% When all the actions in the tree are simulated, \texttt{Rank simulation Tree()} finds the branch with maximum utility value using the Equation \ref{eq:utility}, and accordingly grounds the truth values of the parameters and assign the optimal state to a set of cooperative individuals. If the maximum utility value is zero, the \texttt{Find Optimal state()} makes the current optimal state infeasible and finds a new optimal state to advance the cooperation.

% When the simulation of the optimal state is done the \texttt{Find Next Action()} finds the first undone action in the optimal state and gives the command to the responsible individual. 

When the acknowledgement of an action execution is available (either successful or failed), the \textit{Update Action-State table} function updates the representation in the Action-State table.
If a human operator performs an action that was not suggested, then the \textit{Task Manager} commands the robot to terminate its current action and to go to its \textit{resting configuration}.
Finally, if the Action-State table does not hold any feasible states or state transitions, the cooperation is failed.

\section{Experimental Evaluation}
\label{sec:Evaluation}
% \textbf{[3-4pages]}
% \textcolor{blue}{-- System set up:  codes, ROS, Operating System, Robot, Smart Watch/Phone.} [0.3 page]

\subsection{Implementation and Experiments}
\label{sec:implementation_experiments}

\textsc{FlexHRC+} has been validated using a dual-arm Baxter manipulator, equipped with an additional RGB-D device located on the robot's \textit{head}, and pointing downward to a table where objects to manipulate are located.
In order to gather information about the motion of human operators, we use an LG G watch R (W110) smartwatch for acquiring inertial data from their right wrist.
Data are routed via a Bluetooth connection to an LG G3 smartphone, and then to a workstation through standard WiFi.
The workstation has an Intel Core i7-4790 3.60 GHz $\times$ 8 CPUs, and 16 GB of RAM.
It runs an Ubuntu 14.04 LTS 64 bit operating system.
The architecture has been developed in C++/Python under the ROS Indigo framework, while the smartphone and smartwatch apps have been developed in Java.

% \subsection{Scenario} [1-2 pages]
Three types of experiments have been performed:
(i) assembly operations whereby each action is carried out by the robot or the human operator, to show how the FOL-based representation outperforms standard AND/OR graphs, and the hierarchical AND/OR graph surpasses the single layer counterpart; 
(ii) assembly operations carried out by the robot alone to show the scalability features of the \textit{Task Representation} module and to compare the benefits of the hierarchical AND/OR graph structure in real-world scenarios;
(iii) cooperative assembly operations to show the flexibility, the proactive decision making and the reactive adaptation capabilities of \textsc{FlexHRC+}.
% The AND/OR graph \textit{Task Representation} method is accompanied by an open source software\footnote{Please refer to: \url{https://github.com/kouroshD/ANDOR.git}.}. 
The AND/OR graph representation is accompanied by an open source implementation\footnote{Web: \url{https://github.com/TheEngineRoom-UniGe/ANDOR}.}.

\subsection{Performance Evaluation of the Task Representation Module}
\label{sec:PerformaceComparison}

\begin{figure*}[!t]
\centering
\includegraphics[width=0.98\textwidth]{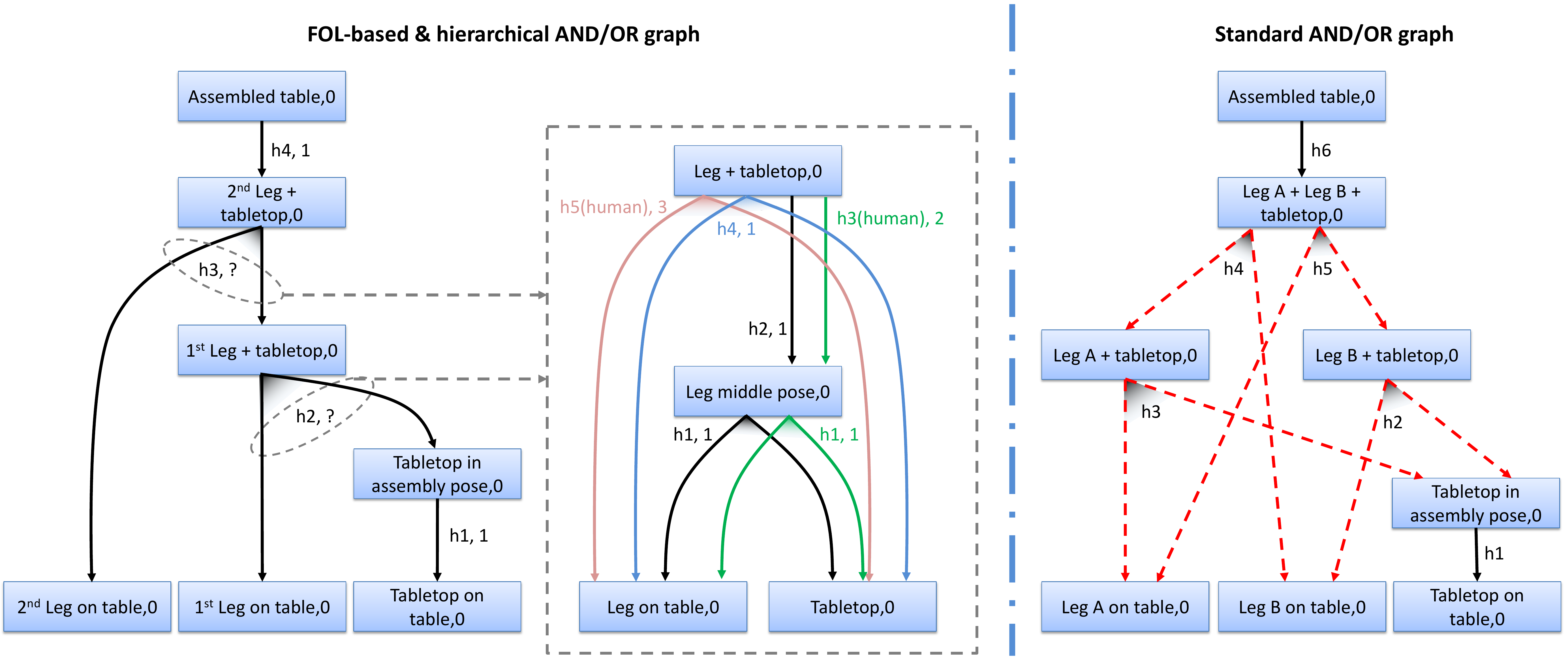}
\caption{An AND/OR graph modelling a table assembly task with two legs. On the left: the high-level AND/OR graph for table assembly. In the middle: the low-level AND/OR graph connecting one leg to a tabletop. On the right: standard AND/OR graph with two identical legs $A$ and $B$. For the sake of clarity, not all the hyper-arcs are shown, i.e., the red dashed hyper-arcs are simplified.}
\label{fig:hierarchical_andOr}
\end{figure*}

In order to carry out a performance evaluation of standard, FOL-based, and hierarchical AND/OR graph representations, let us consider the problem of assembling a table with a different number of legs, whereby we gradually increase the number of table legs to be assembled from one to nine. 
This implicitly makes the graph more complex due to the increasing number of required hyper-arcs.
Figure \ref{fig:hierarchical_andOr} shows the hierarchical (on the left) and the standard version of the AND/OR graph (on the right) for a table assembly with only two legs.
The task is obviously unrealistic, but it can be used to highlight the difference between the single-layer and the hierarchical representations, as well as standard and FOL-based models.
The hierarchical representation is modelled by a FOL-based representation.
For this task, we use a two-layer hierarchical AND/OR graph.
% For this task, we use two layers of hierarchy.

\begin{figure*}[!t]
\centering
\includegraphics[width=0.99\textwidth]{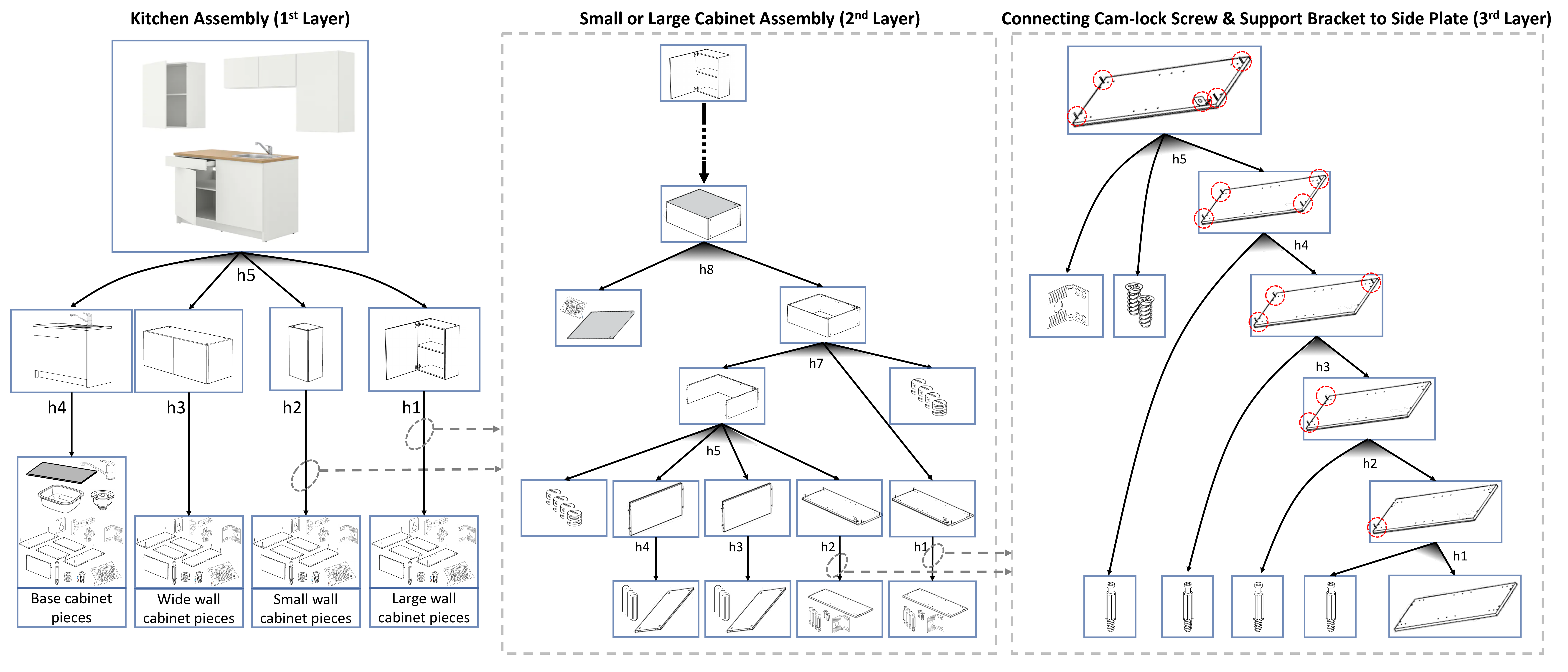}
\caption{A formalisation of an IKEA kitchen assembly task representation using a hierarchical AND/OR graph with five layers according to the products documentation. On the left: the first layer of the AND/OR graph showing high-level kitchen assembly instructions. In the middle: the second layer related to the assembly of the wall cabinet with small or large sizes (second layer). On the right: the third layer with the connection of cam-lock screws and support brackets to the side plates of the cabinets (images courtesy of IKEA).}
\label{fig:kitchen_assembly}
\end{figure*}

% For the sake of clarity

In order to connect a leg to the tabletop according to Figure \ref{fig:hierarchical_andOr} in the middle, there are four cooperation paths that can be followed:
(i) the robot connects the leg and the tabletop directly (\textit{blue} path in the Figure, cost set to $1$); 
(ii) the human operator connects the leg and the tabletop directly (\textit{red} path, cost equal to $2$); 
(iii) the robot places the leg in a new position in the workspace, and later connects the leg to the tabletop (\textit{black} path, cost set to $2$); 
(iv) when the leg is in a new position, the human operator connects it to the tabletop (\textit{green} path, cost equal to $3$). 
In the third case, the reason for introducing a temporary new position for the table leg is that, when a dual-arm robot is used, there might be situations whereby one arm can reach the initial leg position, but only the other one can move it to its final position.
Although the cost of the red and green paths are equal (both of them are $3$), if the robot can perform hyper-arc \textit{h1}, it follows the green path. 
In such a path, the first feasible hyper-arc \textit{h1} is common to the black and green paths.
Therefore, solving this hyper-arc can lead the cooperation to move to the black path, which is characterised by a lower cost (equal to $2$), as well as to the green path. 
Before solving the \textit{Leg middle pose} node,  the robot cannot ascertain whether it can perform hyper-arc \textit{h2} or not, and therefore the green path is followed if possible.
While the cooperation process unfolds, human operators can switch the cooperation to the red path if they deem it fit.

In order to compare standard and FOL-based AND/OR graph representations, let us label the identical legs with $A$ and $B$. 
Even if the objects are identical, their locations in the robot workspace are different. 
Therefore, during execution, the predicate \textsc{Leg(?x)} should be grounded to either $A$ or $B$, so that the leg's location is known to the robot when issuing motion commands.
However, at the representation level, since the legs are identical, the one actually chosen is irrelevant. 
Figure \ref{fig:hierarchical_andOr} on the right hand side shows abstractly all the possible ways to place the two legs on the tabletop when they are grounded in the \textit{Task Representation} module. 
For the sake of clarity, a simplified AND/OR graph is shown in this Figure.
In the experiments, all the details (similar to what is depicted in Figure \ref{fig:hierarchical_andOr} in the middle) are modelled in the same layer. 

\begin{figure}[!t]
\centering
\includegraphics[width=0.97\columnwidth]{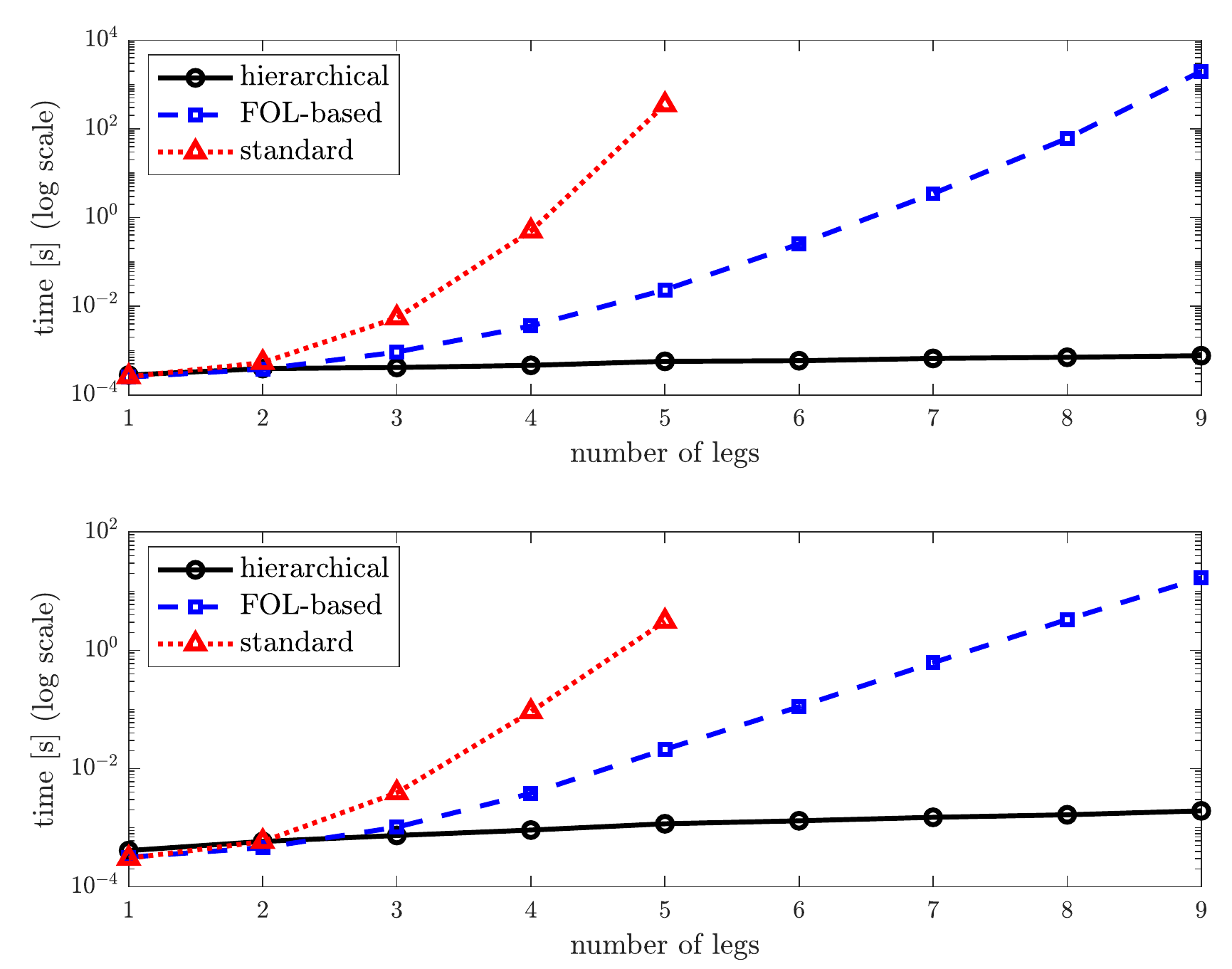}
%logFigure2
%hierarchical_Standard.eps
\caption{The mean computational time in logarithmic scale to solve a single-layer standard (dotted red line), FOL-based (dashed blue line), and hierarchical (solid black line) AND/OR graph for a table assembly task with a varying number of legs: on the top, the offline phase; on the bottom, the online phase.}
\label{fig:andor_efficiency}
\end{figure}
Figure \ref{fig:andor_efficiency} depicts the average computational time (in a logarithmic scale) for the table assembly task in the offline (on the top in the Figure) and online (bottom) phases. 
For each number of considered legs, we have performed the task ten times and we report average timings. 
% \textcolor{blue}{ 
In this experiment, the selection of the cooperation path is done randomly to avoid bias.
% }
As it can be noted, the single-layer AND/OR graph representations, i.e., the standard and the FOL-based graphs, seem to be characterised by an exponential complexity in the number of legs (both offline and online), whereas in the hierarchical representation the computational time grows linearly. 
In the FOL-based, single-layer representation, the average computational time increases from $(2.5 \pm 0.1) \times 10^{-4}$ $s$ (one leg) to $(1.98 \pm 0.02) \times 10^{3}$ $s$ (nine legs) for the offline phase, and from $(3.2 \pm 0.2) \times 10^{-4}$ $s$ to $(1.7 \pm 0.1) \times 10^{1}$ $s$ for the online phase.
In the FOL-based hierarchical case, the average computational time for the offline phase increases from $(2.8 \pm 0.1) \times 10^{-4}$ $s$ (one leg) to $(7.6 \pm 0.3) \times 10^{-4}$ $s$ (nine legs), and for the online phase it raises from $(4.1 \pm 0.2) \times 10^{-4}$ $s$ to $(1.9 \pm 0.1) \times 10^{-3}$ $s$.
As demonstrated in Figure \ref{fig:andor_efficiency}, the FOL-based representation outperforms the standard AND/OR graph, in terms of computational time for both the offline and online phases.
The offline phase takes $(3.49 \pm 0.03) \times 10^{2}$ $s$ to solve the standard AND/OR graph with five legs, and  $(2.33 \pm 0.03) \times 10^{-2}$ $s$ for the equivalent FOL-based graph.
For the online phase, the computational time for the standard AND/OR graph is $(3.1 \pm 0.2) \times 10^{0}$ $s$, whereas for the FOL-based AND/OR graph is $(2.1 \pm 0.1) \times 10^{-2}$ $s$, i.e., a high advantage is given in terms of average time. 
The computational times in case of the single leg case are similar for FOL-based and standard AND/OR graph.

It can be observed, as also shown in Figure \ref{fig:hierarchical_andOr}, that in principle the hierarchical AND/OR graph representation should allow designers to \textit{easily} represent a table assembly task with a varying number of legs without the need for detailing which specific leg a robot or a human operator should connect to the tabletop.
This is characterised by an obvious advantage in terms of representation scalability for increasingly more complex tasks.
In practice, the designer of this particular cooperation process has specified $25$ nodes and $16$ hyper-arcs to model the table assembly with nine legs employing a hierarchical representation, whereas in case of FOL-based representation, $30$ nodes and $47$ hyper-arcs have been identified.
Similarly, to compare standard and FOL-based AND/OR graphs, the table assembly with five legs needs $119$ and $18$ nodes, and $402$ and $27$ hyper-arcs, respectively. 
Clearly, it is not doable in practice to model the assemblage of a table with a high number of legs in case of a standard AND/OR graph because of the modelling complexity. 
Therefore, we can reasonably argue that the adoption of a FOL-based and hierarchical AND/OR graph facilitates to a great extent the modelling process.
Obviously, such an evaluation should be carried by means of a methodological and experimental study, which for its extensive nature is out of scope here.

\subsection{An IKEA Kitchen Assembly Scenario}
\label{sec:IKEA}

Figure \ref{fig:kitchen_assembly} shows the hierarchical AND/OR graph task representation realization for an IKEA kitchen composed of two wall cabinets (small and large), one wide wall cabinet, and a base cabinet with a sink and a faucet. 
The AND/OR graph is implemented according to the assembly documentation of the products provided on the IKEA website \cite{IKEA}.
In this experiment, we benefit from a hierarchical AND/OR graph with five layers, however, only three layers are shown in Figure \ref{fig:kitchen_assembly}.
The interested reader can find the complete AND/OR model in the GitHub repository associated with the developed software architecture.
The kitchen assembly is comprised of $378$ pieces and connectors (such as screws and nuts, but excluding the nails) in total, assembled by $32$ distinct AND/OR graphs used in different layers, spawning $1068$ nodes and $483$ hyper-arcs online.
Exploiting the hierarchical representation, the designer has defined a lower number of nodes and hyper-arcs in the modelling process, i.e., $508$ nodes and $215$ hyper-arcs.
Over ten trials, and by solving hyper-arcs and nodes randomly, the offline and online average computational times are $(5.1 \pm 0.2) \times 10^{-3}$ $s$ and $(1.46 \pm 0.05) \times 10^{-1}$ $s$, respectively.

%%%%%%%  kitchen assembly computational time (sec)
% offline: mean: 5.0807e-03 std: 1.6491e-04
% online: mean: 1.4581e-01  std: 4.9960e-03

%  base cabinet pieces =8+13+6+25+33*4+3 = 187
%  double Door Cabinet=7+15*4+2= 69
% 2* wall cabinet= 2*(7+13*4+2)=122 
%%%%%
%  No of offline written nodes=[113 15 14 9 41 10 9 7 7 46, 9 13 19 11 5 15 11 11 5, 9 10 7 26 7 11 11 9 17 11 7 7 6] => sum=508
%  No of offliine written hyper-arcs =[45 6 5 5 14 5 2 5 5 16, 5 6 8 5 2 8 6 5 2, 4 6 2 9 3 5 5 4 8 5 3 4 2] ==> sum= 215
% %%%% for all the online and/or graphs
% onlineNodes= [ 19 19 11 11 15 15 5 5 11 11 13 9 41 19 19 11 11 15 15 5 5 11 11 13 9 41 19 19 11 11 15 15 5 5  5 11 13 13 9 9 46 7 7 9 9 11 11 11 10 6 7 7 11 7 7 11 10 11 11 17 7 11 7  11 11 13 13 9 9 7 9 9 26 113 14 10 9  15 9] ==> sum= 1068
% onlineHA=       [ 8  8  6  6  8  8  2 2 5  5  6  5 14 8  8  6  6  8  8  2 2  5  5 6  5 14 8  8  6  6  8  8  2 2  2 5  6  6  5 5 16 5 5 4 4 5  5  6  6  2 4 4 6  4 4 6  6  5  5  8  3  5 2  5  5  6  6  5 5 3 4 4 9  45  5  5  2  6  5] --> sum =483

While designing the associated AND/OR graph, we assume that the agents assembling the kitchen can perform the following actions, namely: \textit{keeping a piece}, \textit{approaching a desired pose}, \textit{transporting an object to a desired pose}, \textit{fitting a piece to another piece}, \textit{screwing using bolt and nuts}, \textit{hammering nails}, \textit{following a trajectory with the contact force}, \textit{grasping}, and \textit{ungrasping}. Moreover, the agents should perceive their own actions as well as actions by other agents, recognize the pieces and their features such as the size.

The first layer of the AND/OR graph, as represented in Figure \ref{fig:kitchen_assembly} on the left hand side, decomposes the kitchen assembly task into assembly scenarios for different cabinets.
The assembly of each cabinet is modelled in more detail in lower-level AND/OR graphs, therefore increasing the modularity and scalability at the representation level.
The three cabinets to be attached to the wall are  simpler than the base cabinet, therefore their assemblies are modelled by two additional lower-level layers, whereas the base cabinet assembly (including faucet, sink, strainer, and a drawer) is more complex, and therefore it has been modelled by four extra layers.
In this scenario, the large and small wall cabinets (hyper-arcs \textit{h1} and \textit{h2} of the first-layer AND/OR graph) are similar in shape and are different in size.
Owing to the FOL-based AND/OR graph representation, we can model the assembly of the two cabinets using a unique AND/OR graph.
As shown in the middle of Figure \ref{fig:kitchen_assembly} related to the wall cabinet assembly process, hyper-arcs \textit{h1}, \textit{h2}, \textit{h3}, and \textit{h4} model the connection of different screws, support brackets, or dowels to the cabinet's wooden plank parts.
Hyper-arcs \textit{h5} and \textit{h6} model the connection of different planks using cam-lock nuts for constructing the structure of the cabinet.
Hyper-arc \textit{h8} hammers the backplate of the cabinet to the cabinet structure using nails.
Later, the assembly proceeds with fixing the hinges to the structure and front door, attaching the structure to the wall, and connecting the front door to the structure. 
Finally, the right hand side of Figure \ref{fig:kitchen_assembly} details the hyper-arcs \textit{h1} and \textit{h2} of the cabinet assembly. 
It models the connection of the cam-lock screws and bracket supports to the cabinet side wooden planks. Accordingly, the \textit{Task Manager} specifies the actions to carry out in order to execute hyper-arcs. 

The IKEA kitchen assembly scenario demonstrates the scalability feature of AND/OR graphs with several layers, i.e., the possibility to define layers of different semantics, and to exploit the same lower-level assembly scenario in several higher-level assembly processes.
Although this representation model is here presented only \textit{as it is}, i.e., we do not show a real robot actually collaborating with a human operator in the process, nonetheless we believe it is a good example of a real-world, yet complex assembly process, which could be an excellent use case.

\subsection{A Human-Robot Cooperative Table Assembly Task}
\label{sec:TableAssembly}
% [1-2 pages]

% \textcolor{blue}{Describe the HRC experiments and the goals of the experiments}
% - a photo of the two different tables

\begin{figure}[!t]
\centering
\includegraphics[width=0.95\columnwidth]{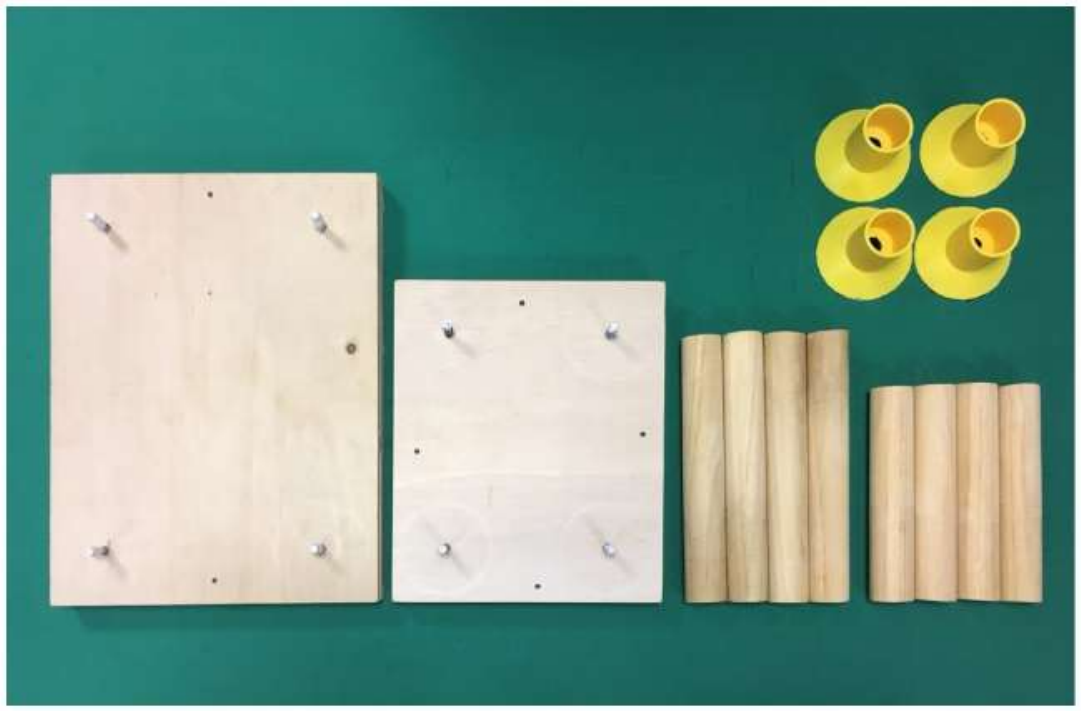}
\caption{The image shows the two different tabletops (left), the two set of legs (bottom-right) and the four 3D printed skirts (top-right).}
\label{fig:tables}
\end{figure}
In a third set of tests, a number of human-robot cooperative table assembly tasks have been carried out using tables different in terms of tabletop sizes and leg lengths, without any change at the task representation level.

These tests demonstrate \textsc{FlexHRC+} capabilities at two levels.
On the one hand, at the \textit{team} level, the human operator and the robot perform the same table assembly task using different cooperation paths online, including no cooperation at all, i.e., the operator or the robot manage the assembly process on their own.
On the other hand,  at the \textit{task} level, the human operator and the robot can cooperate on similar tasks, such as the assembly of tables with different physical properties, without the need for \textit{ad hoc} representations.
In particular, we have used two rectangular tabletops, two sets of four legs, and four customised 3D printed skirts to guide the legs when placing them into the screws for precision compensation, and also to fix the legs to the tabletop, all of them shown in Figure \ref{fig:tables}. 
Hence, four different types of tables can be assembled, which are not \textit{a priori} known to the human operator nor the robot. 
Initially, the legs and the tabletop are randomly located in the shared human-robot workspace. 

%In this experiment, we show how the proactive decision making and reactive adaptation increase the naturalness of the interactions and the robustness to failures.

%The idea behind this assumption is the customized manufactories, where a customer orders a customized product, and therefore the human and robot should produce it cooperatively.
%Moreover, the human is free to decide online how much he wants to participate in the cooperation task, ranging from fully done by the human to fully done by the robot. 
%In these experiments, while the robot is manipulating the object, the human is standing in front of it and monitors the robot actions. In a real assembly line scenario, the human would monitor the task execution of \textit{several} robots, moving between them and intervening whenever necessary. 

Modelled human operator actions include
\textit{pick up} (the human operator picks up one of the legs for manipulation purposes),
\textit{put down} (the manipulated object is put on the table in front of the robot),
and \textit{screw} (the human operator fixes the leg to the tabletop using a rotation movement), whereas robot actions include
\textit{approach} (the robot approaches the grasping pose of an object with one of its end-effectors),
\textit{transport} (the robot moves an object to a desired goal position on the table)
\textit{screw} (similar to the one to be executed by human operators),
\textit{unscrew} (the robot counter-rotates a leg with respect to the screw action),
\textit{grasp} (the robot closes one of its grippers after approaching an object),
\textit{ungrasp} (the robot opens a gripper), 
and \textit{update workspace} (the robot updates its internal representation of the workspace using perception modules).

We model the cooperative table assembly task with four legs using a single-layer FOL-based AND/OR graph similar to the one shown in Figure \ref{fig:hierarchical_andOr}. 
The AND/OR graph representation encompasses all the possibilities, ranging from the case whereby either the human operator or the robot assemble the whole table on their own, to various cases in which the operator and the robot cooperate. 

%We reach these flexibilities through the proactive decision making and reactive adaptation of the \textit{Task Manager}, the rich cooperation models of the \textit{Task Representation}, the online object recognition and automatic computation of different object frames, and finally the action modules.

%The architecture manages different objects and poses automatically and supports the robot capability to manipulate the objects and cooperate with the human in a partially-structured environment. Furthermore, if the human changes the objects pose while cooperating, the robot updates its representation of the workspace and decides online how to continue with the cooperation.

The table assembly task has been performed eight times as presented in the accompanying video\footnote{Web: \url{https://youtu.be/CEIARyW422o}.}.
Results are summarised in Table \ref{tab:userTest}.
The Table shows that the overall execution of \textit{Task Manager}, \textit{Task Representation}, and \textit{Simulator} require less than $2 \%$ of the complete table assembly task, with a low standard deviation. 
The more time consuming portions of the cooperation are the actions performed by the human operator or the robot. 
The high standard deviation associated with experiment lengths is mainly due to human decisions taken online, and to the significant difference between the human and robot speeds in performing the actions. 
As a reference, the maximum robot joint angular velocity, and end-effector angular and linear velocities are $0.6$ $rad/s$, $0.8$ $rad/s$, and $0.4$ $m/s$, respectively.
\begin{table}[!t]
\caption{Computational performance of \textsc{FlexHRC+} modules for the table assembly task.}
\begin{center}
\begin{tabular}{@{}C{2.7cm}C{1.4cm}C{1.6cm}C{1.0cm}@{}}
\toprule
\textit{Computation}		& \textit{avg time [s]}	& \textit{avg time [\%]}	& \textit{std [s]}\\
\hline
%\\
\textit{Task Manager} 			& 0.01			& 0.00			&  0.00 \\
\textit{Task Representation} 	& 0.57 			& 0.23			& 0.01 \\
\textit{Simulator} 				& 4.15 			& 1.66			& 0.40 \\
\textit{Robot actions} 			& 219.45 		& 87.75			& 56.90 \\
\textit{Human actions} 			& 25.82 		& 10.32			& 5.67 \\
\textit{Total}					& 250.09		& 100.00		& 51.87 \\
\bottomrule
\end{tabular}
\end{center}
\label{tab:userTest}
\end{table}
To avoid repetitions in showing the results for different parts of the experiments, we divide the cooperative table assembly task in three segments: 
in the first (tabletop placement), the robot places the tabletop in the workspace; 
the second segment (legs fix) is related to all operations to fix the legs to the tabletop; 
in the last segment (check) the human operator checks all connections.

\begin{figure}[!t]
\centering
\includegraphics[width=0.95\columnwidth]{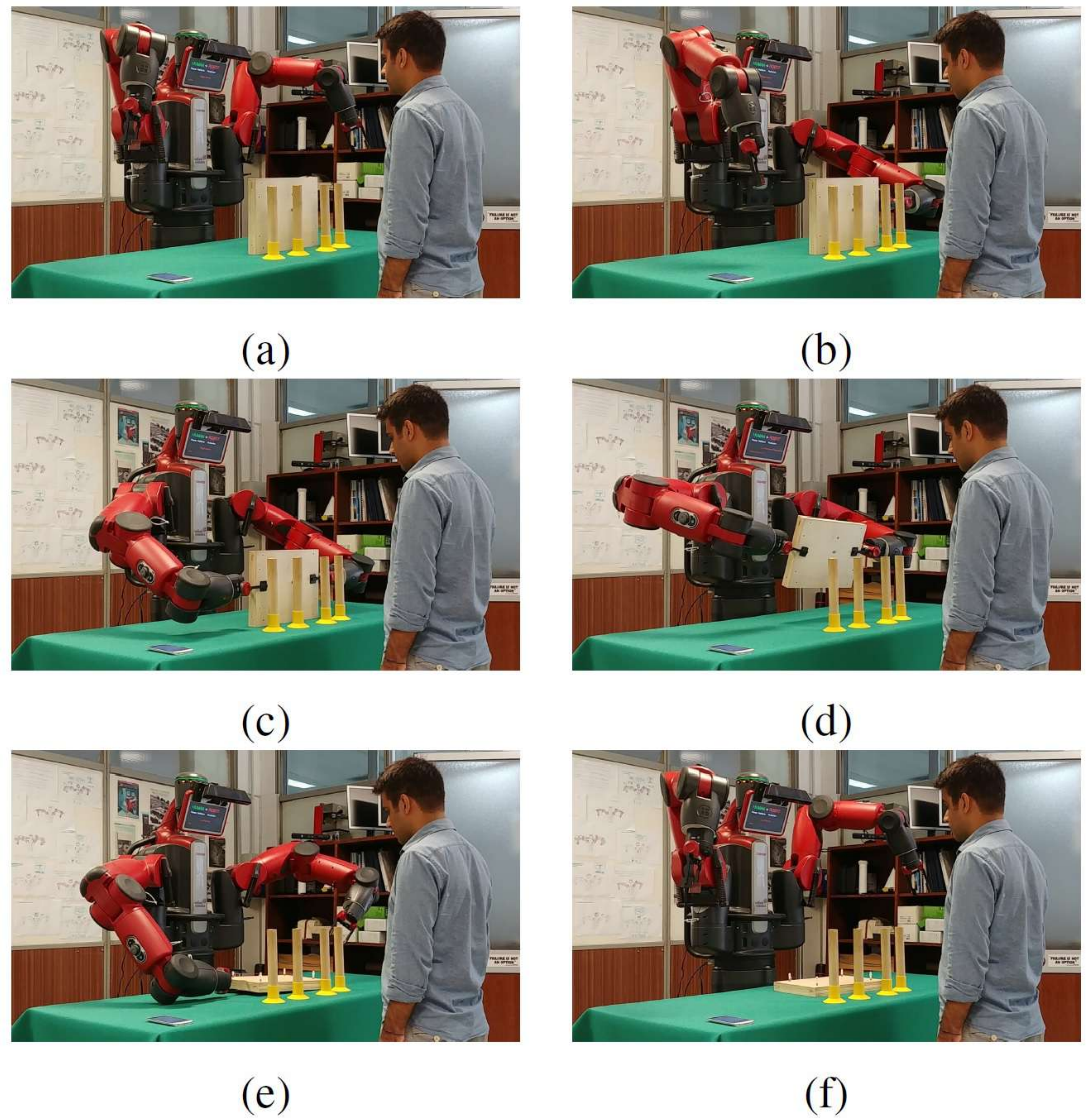}
\caption{The sequence of actions associated with tabletop placement by the robot.}
\label{fig:tabletopPlacement}
\end{figure}
Figure \ref{fig:tabletopPlacement} shows an example of  tabletop placement. 
In our experiments, this sequence is always carried out by the robot: 
%and the robot can manage the task for different types of tabletops or in different poses autonomously. 
the robot's left and right arms approach the tabletop (a-b), grasp it (c), change its orientation and place it on the table horizontally (d-e), and finally return to the resting pose (f).

\begin{figure*}[!t]
\centering
\includegraphics[width=0.95\textwidth]{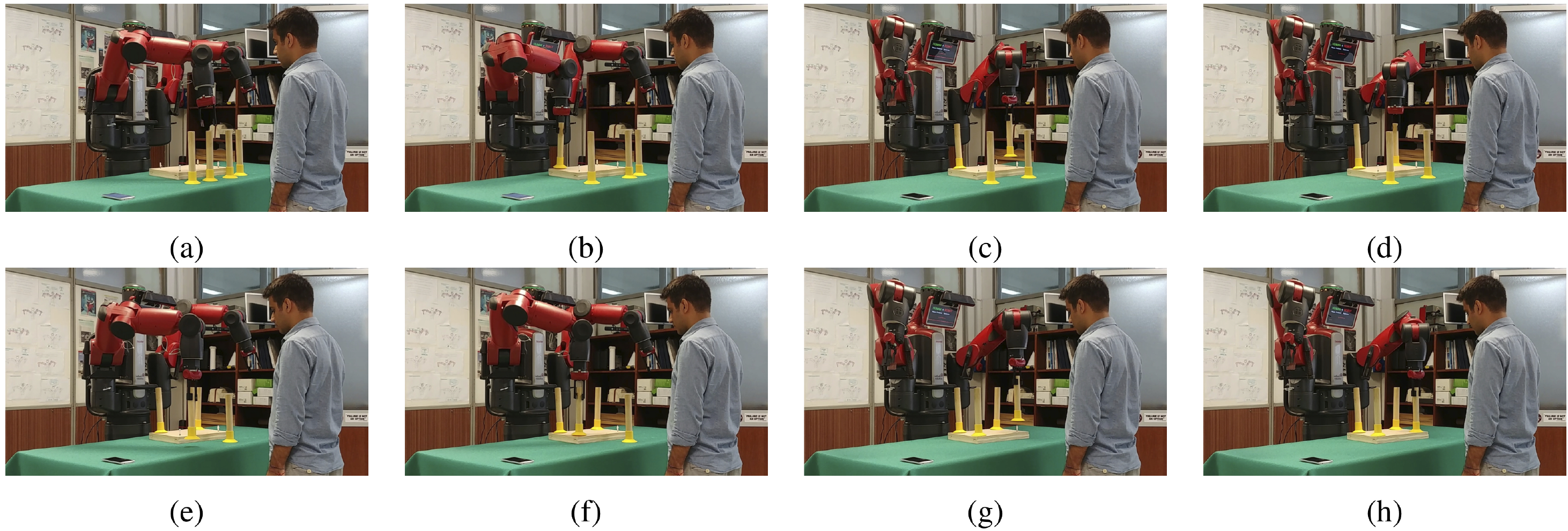}
\caption{The robot connects all the legs to the tabletop.}
\label{fig:AllRobot}
\end{figure*}

\begin{figure*}[!t]
\centering
\includegraphics[width=0.95\textwidth]{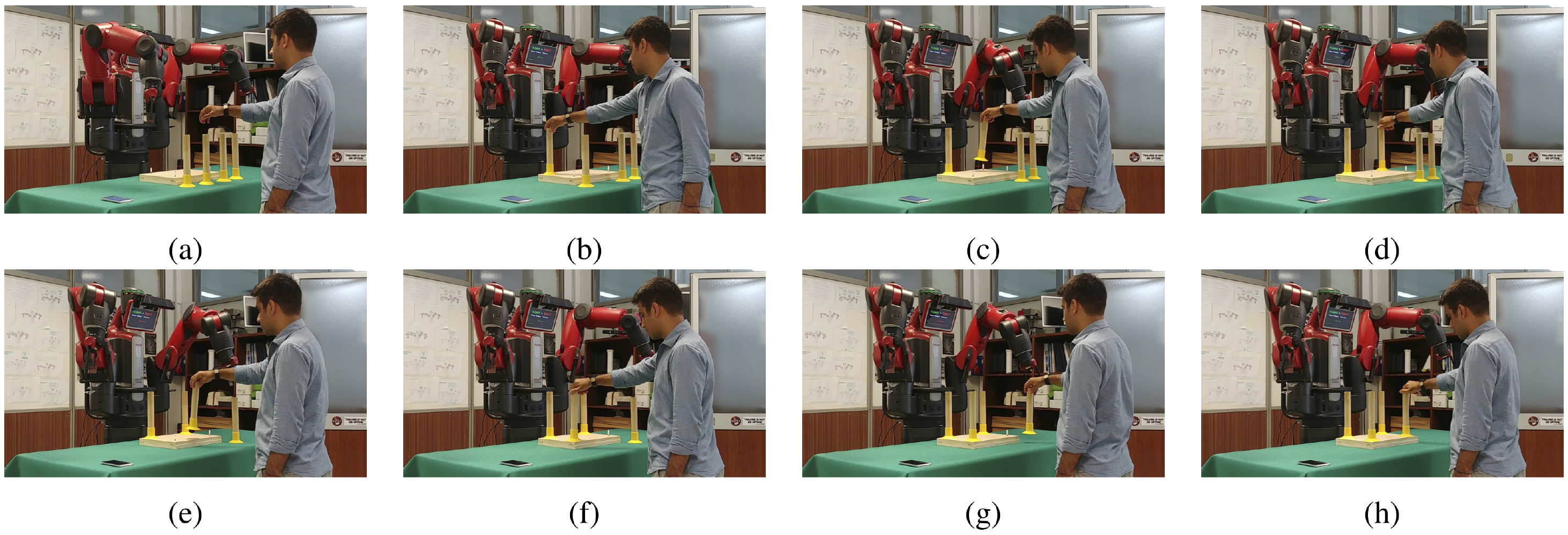}
\caption{The human operator connects all the legs to the tabletop; the robot adapts to human decisions online.}
\label{fig:AllHuman}
\end{figure*}

\begin{figure*}[!t]
\centering
\includegraphics[width=0.95\textwidth]{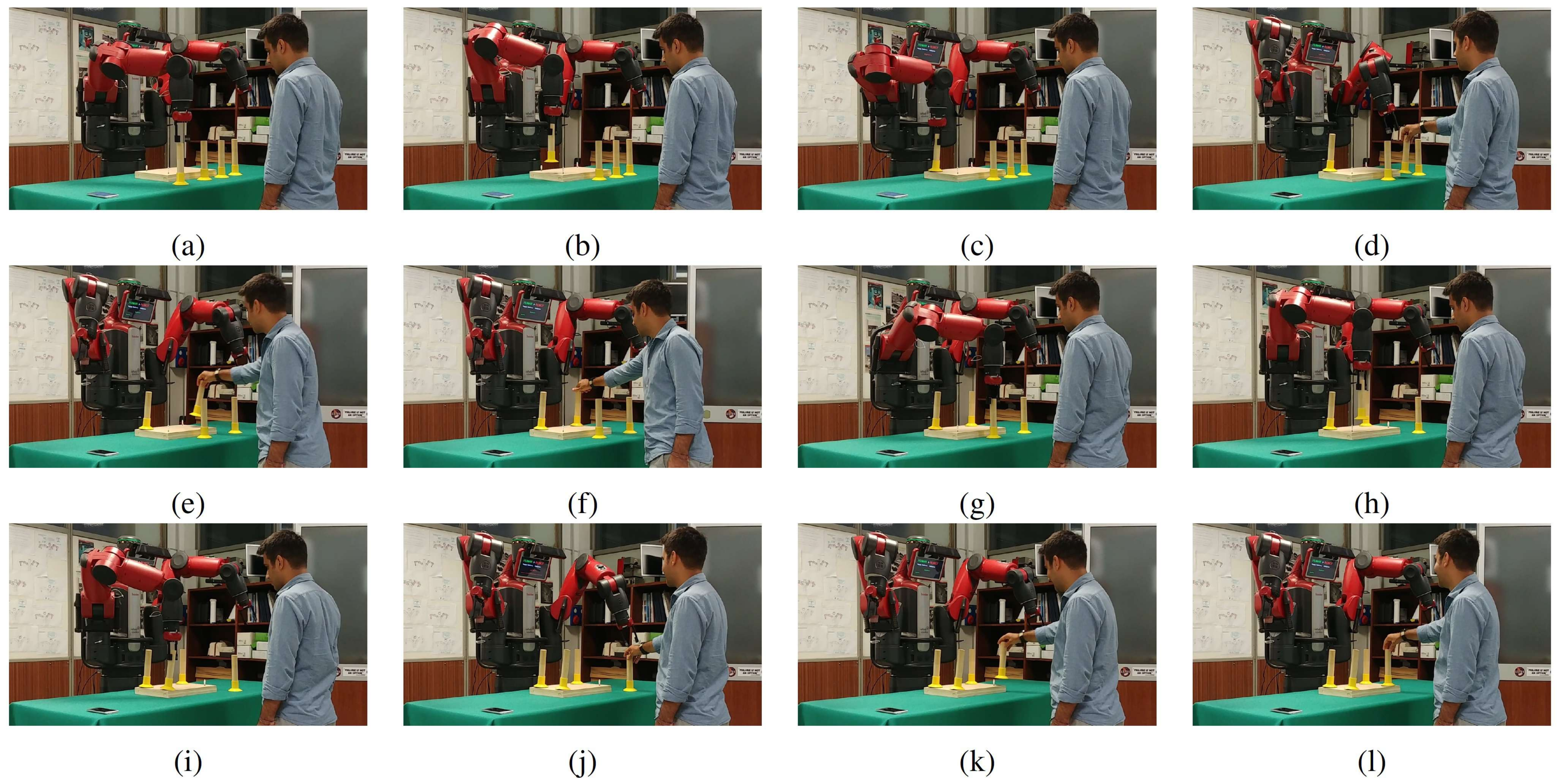}
\caption{The human operator and the robot connect the legs to the tabletop cooperatively; the robot adapts in any case to human decisions.}
\label{fig:ReactiveAdaptation}
\end{figure*}

\begin{figure*}[!t]
\centering
\includegraphics[width=0.95\textwidth]{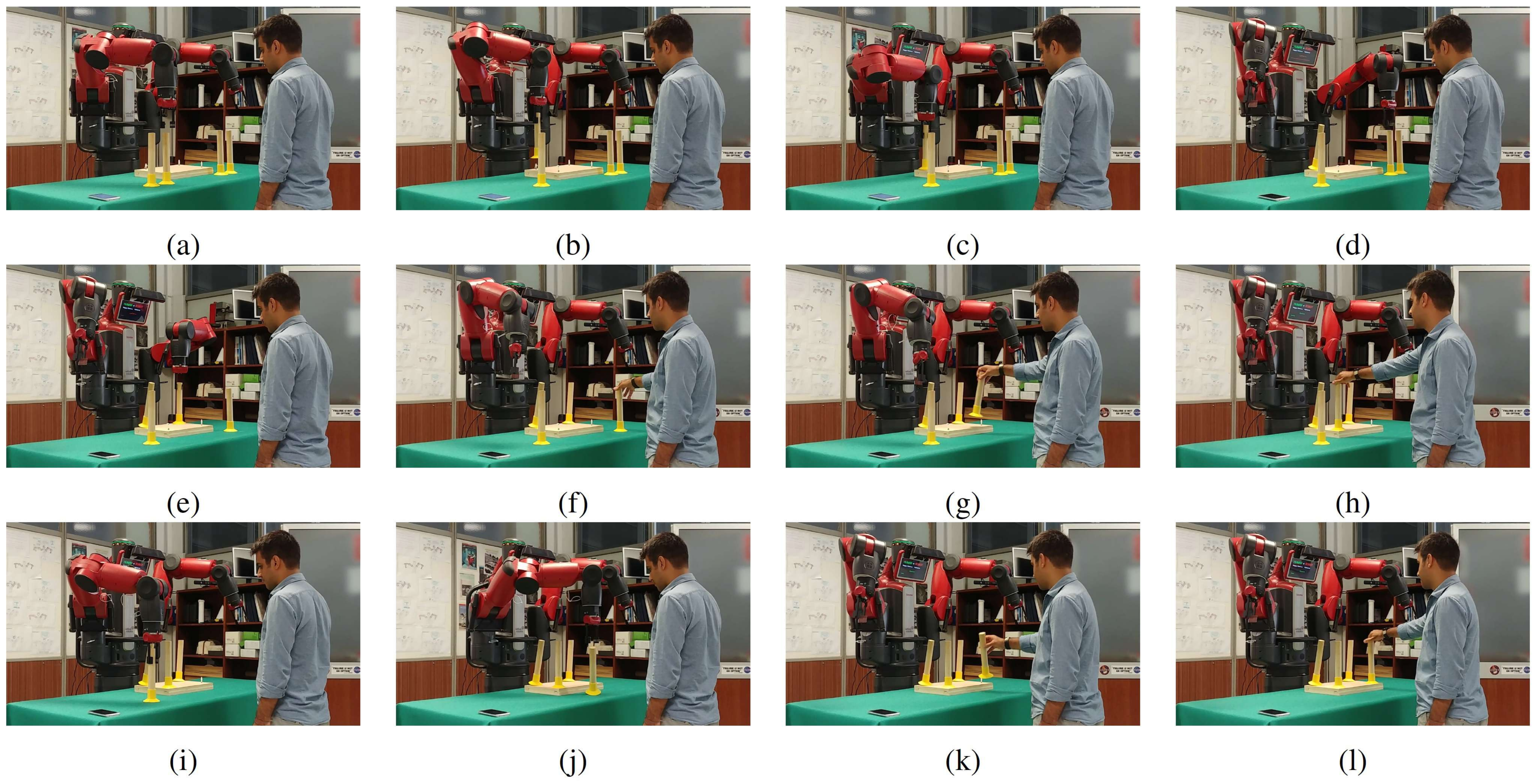}
\caption{The human operator and the robot connect the legs to the tabletop cooperatively; the robot adapts to human decision, and asks proactively to the human to perform a certain task.}
\label{fig:ProactiveAdaptation}
\end{figure*}

Figures \ref{fig:AllRobot}, \ref{fig:AllHuman}, \ref{fig:ReactiveAdaptation}, \ref{fig:ProactiveAdaptation} show different situations whereby the human operator and the robot perform a legs fix cooperatively. 
%Depending on the situation, the human operator and the robot define the amount of the cooperation online. 
They are free to choose which leg to pick up for manipulation, whereas the order of the bolts on the tabletop is \textit{a priori} defined
in our scenario.
Figure \ref{fig:AllRobot} shows the case in which the robot connects all the legs to the tabletop. 
At the beginning, the robot can choose which one of the four legs to pick up, as well as which arm to use to do it.
As shown in the previous Section, this is modelled using FOL-based predicates whose literals must be anchored to actual \textit{percepts}. 
To do so, the robot evaluates the utility function for all eight options (i.e., all the couples of legs and arms)  using the \textit{Simulator} module. 
In this specific example, the second leg from the right and the right arm are selected (a-b, robot point of view).
The human operator decides not to intervene in the assembly process, and the robot performs all the assembly operations autonomously.
Therefore, the robot follows the blue cooperation path shown in Figure \ref{fig:hierarchical_andOr}, and fixes all the legs to the tabletop.
Figure \ref{fig:AllHuman} shows a case in which the human operator decides to connect all the legs to the tabletop.
The robot tries, by default, to follow the minimum cost cooperation path (in blue in Figure \ref{fig:hierarchical_andOr}), but as soon as the \textit{pick up} action by the human operator is detected, the robot follows the corresponding cooperation path in red.
Figure \ref{fig:ReactiveAdaptation} shows the case in which the human operator and the robot cooperate to connect all the legs to the tabletop.
In particular, in (a-c) it is shown how the robot fixes the rightmost leg to the tabletop. 
Later, the robot decides to connect another leg with the left arm, but the human operator intervenes and performs the fix operation to the second tabletop bolt (d-f).
The robot adapts to the human decision when the human action is recognised.
It updates its representation of the workspace via the \textit{Object and Scene Perception} module, it updates the AND/OR graph and determines the new set of feasible states. 
Then, the robot connects the third leg to the tabletop, and finally the human decides to fix the last leg (g-l).
Figure \ref{fig:ProactiveAdaptation} shows how the robot adapts to human decisions, and how it can proactively request the human to perform a certain task, which it cannot perform. 
In (a-e) the robot connects the first two legs to the tabletop.
Afterwards, the human operator decides to connect the third leg to the tabletop (f-h). 
Finally, in (i-l) a situation is depicted whereby the robot is not capable of performing a given task, and therefore it asks the human operator to connect the last leg to the tabletop. 
By following the green cooperation path in Figure \ref{fig:hierarchical_andOr}, first the robot transports the leg in front of the human operator, and finally the operator connects it to the tabletop.

% He decide for the cooperation in between, which connects some of the legs to the tabletop (Figure ...). 
% Moreover, if the robot can not connect one of the legs to the tabletop, it asks the human for cooperation to connect the leg to the tabletop (Figure ...).

\begin{figure*}[!t]
\centering
\includegraphics[width=0.95\textwidth]{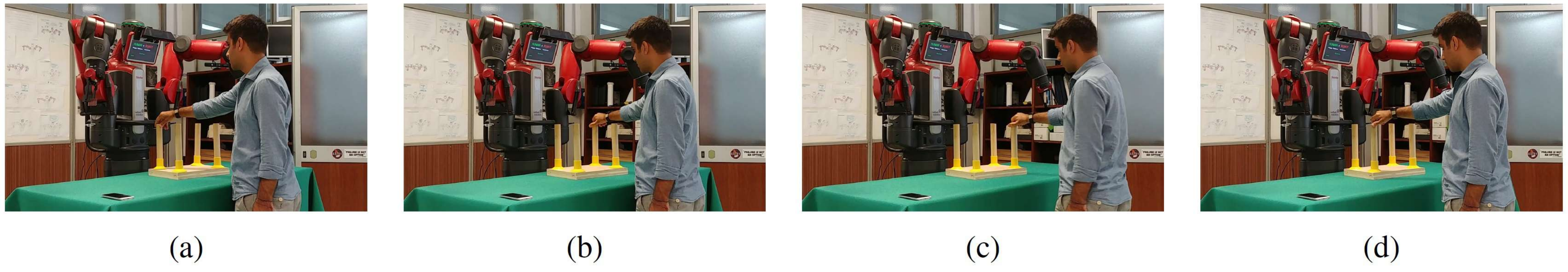}
\caption{A sequence of check actions the human operator carries out to verify all connections between the legs and the tabletop.}
\label{fig:monitoring}
\end{figure*}
Finally, Figure \ref{fig:monitoring} shows the human operator controlling all leg connections to the tabletops at the end of the task.
%If the connections are loose, the human fixes them.

It is noteworthy that in Figures \ref{fig:AllRobot}, \ref{fig:AllHuman}, and \ref{fig:ProactiveAdaptation}, the human operator and the robot assemble the large tabletop with the long legs, while in Figure \ref{fig:ReactiveAdaptation} they assemble the small table with the short legs.
These examples show the ability of \textsc{FlexHRC+} to handle different real-world objects, i.e., anchoring different percepts to the same symbol-mediated predicates in the FOL-based representation.
As a consequence, we can considered the perception layer and the representation layer as decoupled, i.e., changes in the robot workspace need not to be represented.
%This feature not only increases the flexibility in task level but also supports the idea of the scalability.

\subsection{Discussion}
\label{sec:Discussion}

% \textcolor{blue}{R1: effectiveness and performance of human operators ~ robot motion predictability, cobots prevent psychological discomfort, stress, and a high induced cognitive load on human operators }
% Regarding the function requirement $R_1$, in this paper we propose a simulation based decision making approach for the robot. Such results can be used to exhibit to the human the robot motions and actions while start performing them. Displaying these information to the human increases the effectiveness of the collaboration, increases the safety and reduce the cognitive load on the human operators. To increase the naturalness and intuitiveness such a information can be combined with Augmented Reality (AR) technologies.

As posited in the Introduction, \textsc{FlexHRC+} targets three of the identified requirements for HRC, namely 
$R_2$, which is related to the trade-off between flexibility and actions aimed at meeting the cooperation objectives,
$R_4$, whereby the robot should exhibit flexible decision making capabilities, and
$R_5$, the capability of doing so exploiting hierarchical task representation structures.

In the first case, it is posited that a control architecture for collaborative robots should not only provide the necessary autonomy for general-purpose action planning and execution, but should also allow the collaborative robot to adapt to the behaviour of human operators.
Such an adaptation should be limited to the available plans reaching the cooperation goal. 
\textsc{FlexHRC+} copes with $R_2$ by equipping the robot with the ability of autonomously scheduling the most appropriate actions to carry out, while exploiting perception and action modules to adapt to human operators. 
The hierarchical, FOL-based AND/OR graph representation is capable of switching among possible, intrinsically feasible plans, and among them following the currently optimal one. 
This is achieved also by means of an in-the-loop, simulation-based approach to anchor internal symbols to actual perceptions. 
The results of the simulation include the actions to be performed, as well as the upcoming end-effector trajectories. 
Although human-robot communication is not the focus of the paper, it is noteworthy that current work is devoted to the use of Augmented Reality technology to intuitively communicate upcoming robot actions to human operators \cite{Green2008, Michalos2016}.
In the second case, \textsc{FlexHRC+} enables some level of flexibility as far as the representation of the cooperative task is concerned, as required by $R_4$.
The framework separates low-level perception activities from the high-level structure of the cooperation task. 
%Perception modules are expected to handle uncertainties in sensory data. 
At the task and architectural levels, the robustness to noisy or inaccurate perception is mainly related to what happens in the \textit{Knowledge base} module. 
Therein, two assumptions are made, i.e., closed-world and continuity, which imply that any change in the robot workspace is the consequence of actions performed by the human operator or the robot. 
In the current version of \textsc{FlexHRC+}, all relevant information for robot behaviour is maintained in the knowledge base. 
If newly perceived information is not compatible with stored information, then the knowledge base enters an inconsistent state, and waits for new, compatible, information.
In previous work \cite{Capitanelli2018}, we have shown how to deal with inconsistent states in a knowledge base, by using the notion of normative knowledge (i.e., information expected to hold at a given time instant) to plan for a series of robot actions aimed at solving inconsistencies.
Such an approach is not integrated with \textsc{FlexHRC+} yet.
%Moreover, uncertainties in action execution are handled in the \textit{Robot Execution Manager} and \textit{Controller} modules.
%These modules communicate to the \textit{Task Manager} using high-level information at the semantic level.
%
As far as $R_5$ is concerned, a hierarchical representation is a key enabling factor to model complex scenarios adopting a bottom-up approach and achieving lower reasoning times.
In the \textit{Task Representation} module, the use of hierarchical, FOL-based AND/OR graphs simplifies the definition of qualitatively similar tasks, which can be then used to model different phases of the cooperation, and it greatly increases the computational efficiency in graph traversal, which is of the utmost importance to allow collaborative robots to be reactive and adaptive to plan changes. 
This allows for building a library of simpler tasks, which can be composed together to create composite tasks \cite{MastrogiovanniSgorbissa2013}.
In this respect, current work is devoted to integrate in \textsc{FlexHRC+} the capability of learning from human demonstrations and through interactions with the environment \cite{Carfietal2019}.

\section{Conclusions}
\label{sec:Conclusions}
% \textbf{[0.25-0.5 page]}

In this paper, we have introduced a human-robot cooperation framework enabling a greater flexibility and scalability in assembly tasks.
In terms of flexibility, two levels are involved.
At the task level, the robot deals with the differences in the objects being manipulated, without the need for explicitly encoding such differences in the cooperation model.
At the cooperation level, different modules in the framework allow the human operator or the robot to choose the degree of cooperation as the task progresses.
As far scalability is concerned, it is possible to design complex cooperation scenarios building upon simpler ones.
To this aim, a First Order Logic hierarchical task representation has been developed and integrated within the framework.
%Several experimental results have been presented in this work to show the increased flexibility, scalability and expressiveness of the proposed task representation.  
%Previous papers have focused on evaluating its effectiveness and efficiency \cite{Darvish2018Mechatronics,Darvish2018Roman}.
The introduced FOL-based hierarchical AND/OR graph structure makes task representation efficient enough to perform highly complex scenarios, while at the same time simplifying the design phase, and ensuring explainable robot behaviour. 
However, a designer is still required to manually build the graph.
This implies a deep knowledge of the assembly task as well as of the involved robot capabilities.
Obviously enough, such a process is time-consuming, and it is prone to modelling errors or inaccuracies.

In order to cope with these shortcomings, as mentioned in the Introduction, a must-have feature for next generation collaborative robots would be the capability of learning the task structure both at the task and the action level \cite{Stopp2003, Ekvall2008, Argall2009, Konidaris2012}.%Nikolaidis2015 
While at the task level a robot may learn the sequence of discrete actions \cite{Ekvall2008, Munzer2017}, at the action level it should learn how to control its movements to reach a given goal \cite{Argall2009, kormushev2010}. 
%The robot may learn through the interaction with the world and/or human or through the human demonstration \cite{Argall2009}. 
We plan to tackle such learning aspects in a future development of \textsc{FlexHRC+}.

\bibliography{./bib/bibliografia,./bib/hri,./bib/fw}

\end{document}